\documentclass[]{gtech}

\usepackage{amssymb}
\usepackage{bigdelim}
\usepackage{todonotes}
\usepackage{longtable}
\usepackage{tabularray}
\usepackage[most]{tcolorbox} 
\usepackage{algorithm}  
\usepackage{algpseudocode} 
\usepackage[toc,page]{appendix}
\usepackage[normalem]{ulem}
\usepackage{adjustbox}
\usepackage{natbib}
\usepackage{colortbl}  
\usepackage{subcaption}
\usepackage{pifont}
\usepackage{amsmath}
\usepackage{amsfonts}       
\usepackage{mathrsfs}
\usepackage{tabularx}
\usepackage{hyperref}       
\usepackage{url}            
\usepackage{booktabs}       
\usepackage{nicefrac}       
\usepackage{microtype}      
\usepackage{xcolor}         
\definecolor{ourreward}{HTML}{32bbee}
\definecolor{sparsereward}{HTML}{ed9a81}
\usepackage{graphicx}
\usepackage{tikz}
\usepackage{fontawesome5}
\usepackage{twemojis}
\usepackage{wrapfig}
\usepackage{needspace}
\usepackage{cleveref}
\usepackage{fvextra}
\tcbuselibrary{listings, breakable}

\usepackage{bbding}
\usepackage{perpage}
\MakePerPage{footnote}
\usepackage{caption}
\usepackage{CJK}

\usepackage{bbm}
\useunder{\uline}{\ul}{}

\newcommand{\searchos}{\textbf{SearchOS}}

\title{SearchOS-V1~\raisebox{-0.12\height}{\includegraphics[height=1em]{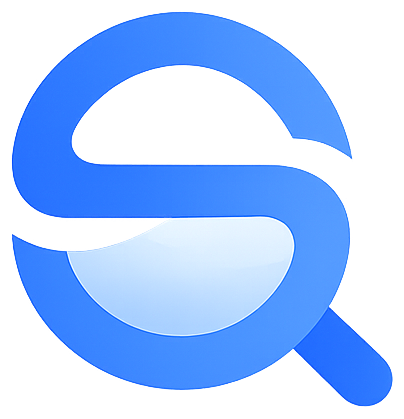}}: Towards Robust Open-Domain Information-Seeking Agent Collaboration}

\author[1,2,*,\ddagger]{Yuyao Zhang}
\author[2,*]{Junjie Gao}
\author[2]{Zhengxian Wu}
\author[2]{Jiaming Fan}
\author[2]{Jin Zhang}
\author[2]{Shihan Ma}
\author[2]{Yao Yao}
\author[2]{Weiran Qi}
\author[2]{Chuyan Jin}
\author[2]{Guiyu Ma}
\author[2]{Xingzhong Xu}
\author[2]{Kai Yang}
\author[1]{Ji-Rong Wen}
\author[1,\dagger]{Zhicheng Dou}

\affiliation[1]{Gaoling School of Artificial Intelligence, Renmin University of China }
\affiliation[2]{Ant Group}

\contribution[*]{Equal contribution}
\contribution[\dagger]{Corresponding author}
\contribution[\ddagger]{Work done during an internship at Ant Group}

\abstract{
\vspace{-1em}
\begin{center}
\bfseries Abstract
\end{center}
\vspace{-0.5em}
Recent advances in Tool-Integrated Large Language Models have made web search a core capability of information-seeking agents. However, as interaction histories grow, agents increasingly struggle to track task progress. When search attempts fail to yield useful evidence, current single- and multi-agent systems can become trapped in repetitive loops, wasting search budgets and ultimately compromising the quality and completeness of the final output. We introduce \searchos{}, a system-level multi-agent framework that turns fragile, implicit search progress into explicit, persistent, and shared state. First, we formulate open-domain information seeking as \emph{relational schema completion with grounded citations}, where agents discover entities, populate attributes across linked tables, and anchor each value to source evidence. Then we design Search-Oriented Context Management (SOCM), which externalizes the evolving state into Frontier Task, an Evidence Graph, a Coverage Map, and Failure Memory. Built on SOCM, \searchos{} applies a pipeline-parallel scheduling mechanism that overlaps the execution of sub-agents and continuously refills freed slots with tasks targeting unresolved coverage gaps to improve utilization and throughput. To schedule and control the execution of search agents, \searchos{} introduces a Search Tool Middleware Harness that intercepts model and tool interactions to record grounded evidence and react to stalls or budget exhaustion, and provides a reusable hierarchical skill system comprising strategy and access skills to augment the agents' search process and avoid repeating failed search patterns across runs. On WideSearch and GISA, \searchos{} leads all metrics among the evaluated single- and multi-agent baselines, paving the way toward robust information-seeking collaboration.
}

\gtechdata[\textcolor{gtechblue}{\faGlobe}\ Website]{\url{https://antins-labs.github.io/SearchOS}}
\gtechdata[\faGithub\ Code]{\url{https://github.com/antins-labs/SearchOS}}
\gtechdata[\textcolor{red}{\faYoutube}\ YouTube]{\url{https://www.youtube.com/playlist?list=PLXMUs3Ayz3EQ}}

\begin{document}
\maketitle
\section{Introduction}
\label{sec:intro}

\begin{figure}[t]
    \centering
    \includegraphics[width=0.9\textwidth]{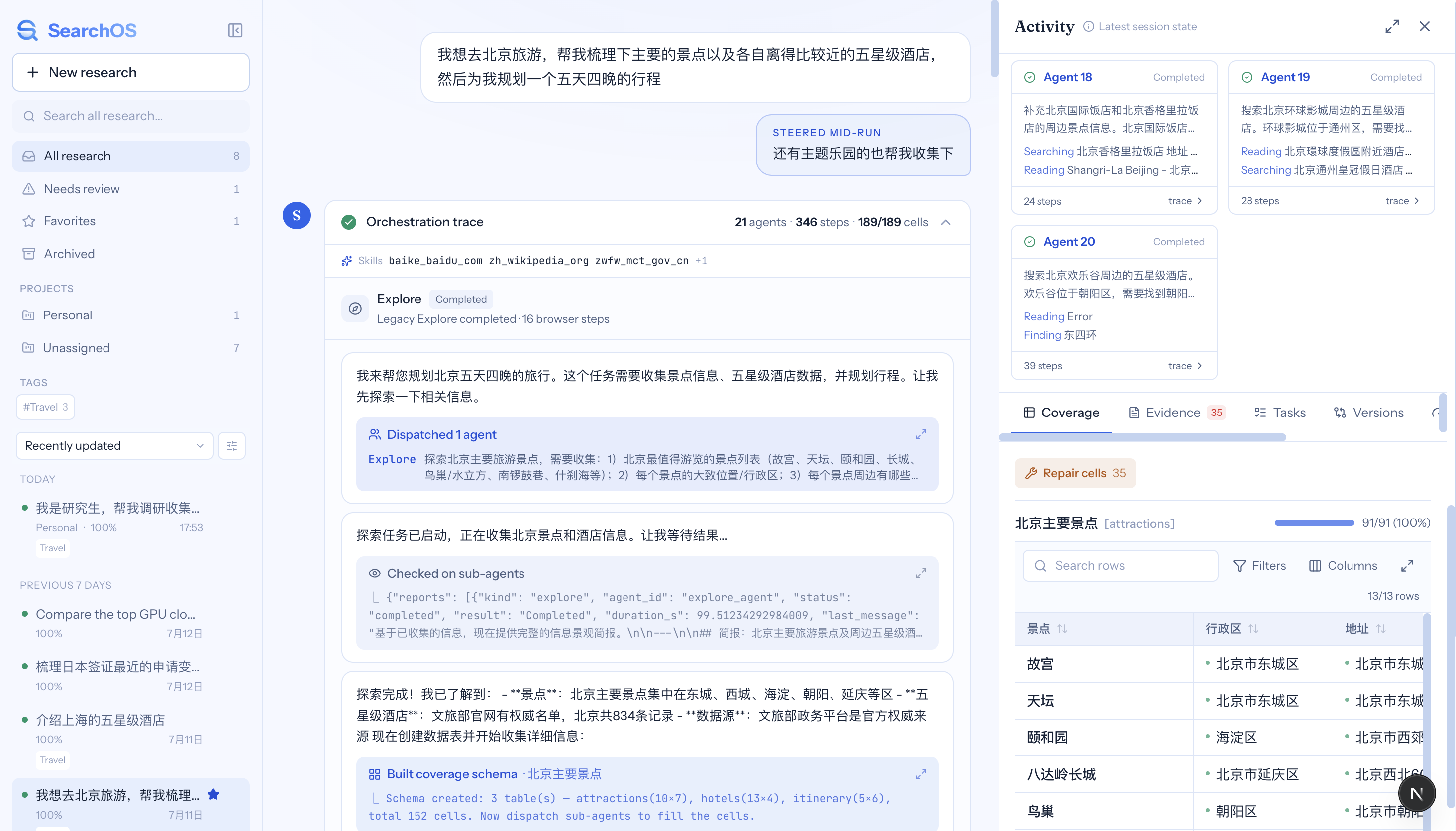}
    \caption{\searchos{} interface for a long-horizon information-seeking task. The workspace exposes the orchestration trace, pipeline parallel agent activity, and relational schema coverage.}
    \label{fig:searchos_preview}
\end{figure}

Recent advances in tool-integrated large language models (LLMs) have made web search a core capability of information-seeking agents~\citep{yao2022react,webgpt,searchr1,webthinker}. These agents can iteratively search, browse, and reason over external sources, extending language models beyond the knowledge available in their parameters. As the field moves toward broad, open-domain and long-horizon tasks~\citep{wong2025widesearch,zhu2026gisa}, an agent must gather large collections of facts, reconcile evidence across sources, and produce a structured answer with verifiable citations.

Stronger search capability alone, however, does not make this process reliable. As interaction histories grow, agents lose track of what has been established and what remains unresolved. Evidence becomes buried in context, making omissions, redundant collection, and conflicting claims difficult to detect. When a search path yields no useful information, a single agent may repeatedly issue similar searches or continue exploring the same dead end, wasting its budget and degrading the final answer. Simply adding more agents does not eliminate these problems: parallel workers may duplicate effort, disagree on the intended fields, or leave execution slots idle while the system waits for the slowest task.

These failures arise because conventional agents treat plans, progress, evidence, and failures as transient conversation content. \textbf{Search state should be maintained by the system rather than inferred repeatedly from interaction history}. A long-horizon search system should expose what entities and attributes have been covered, preserve the provenance of every value, coordinate workers around explicit gaps, and intervene when execution stalls. It should also retain effective search and access patterns across sessions instead of rediscovering them for every task.

Based on this principle, we present \searchos{}, a multi-agent framework for structured and observable open-domain information seeking. We first formulate an information-seeking task as \emph{relational schema completion with grounded citations}. Following relational modeling and normalization principles~\citep{codd1970relational,bernstein1976synthesizing}, a relational search schema may contain multiple tables connected through primary--foreign-key relations. The system jointly discovers the entity rows, fills their attributes, and maintains a citation matrix that maps each value to a source URL and an anchored excerpt. This formulation, related to recent table-completion views of long-horizon information seeking~\citep{tableassearch}, turns an open-ended request into a concrete search target and makes progress measurable at the level of entities and attributes.

We then introduce \textbf{Search-Oriented Context Management} (SOCM) to externalize execution state and share Frontier Task, an Evidence Graph, a Coverage Map, and Failure Memory across all agents. A central orchestrator decomposes the schema into coverage-oriented tasks and coordinates specialized explore, search, and writer agents. Rather than dispatching workers in synchronized batches, \searchos{} uses pipeline-parallel scheduling: different sub-tasks progress concurrently, and each released agent slot is immediately assigned a new unresolved schema gap. This design follows the utilization advantage demonstrated by pipeline parallelism in GPU training~\citep{huang2019gpipe,narayanan2019pipedream}; it reduces straggler-induced idle time and increases search throughput.

Reliable execution also requires controls that should not depend on an agent remembering to invoke them. We therefore introduce a \textbf{Search Tool Middleware Harness} that intercepts model and tool interactions to inject relevant state, extract and anchor evidence, enforce budgets, and detect repeated or stalled trajectories. Agents can focus on local search decisions while the harness maintains global execution invariants. Finally, we design a hierarchical search skill system that separates reusable strategy skills---how to search---from site-specific access skills---how to reach and extract information from a source. The system routes these skills by task and source and refines them from successful and failed trajectories. 

Experiments on WideSearch and GISA show that \searchos{} leads all reported F1 metrics among the evaluated single- and multi-agent baselines. It achieves 80.3 item-level F1 on WideSearch and 76.5 set F1 on GISA, exceeding the strongest baseline on the latter by 13.4 points, paving the way toward robust information-seeking collaboration. Our main contributions are as follows:

\begin{enumerate}
    \item \textbf{Relational Search Formulation.} We formulate open-domain information seeking as relational schema completion with grounded citations, providing a unified and verifiable objective for entity discovery, attribute completion, and evidence attribution.

    \item \textbf{Stateful, Pipeline-Parallel Collaboration.} We introduce SOCM and a continuous orchestration scheme that coordinate specialized agents through shared search state and dynamically assign work around unresolved coverage gaps.

    \item \textbf{Middleware-Governed Search Agent Execution.} We introduce a Search Tool Middleware Harness that moves evidence processing, context management, budget enforcement, and behavioral safeguards outside agent prompts.

    \item \textbf{Hierarchical Search Skills.} We design a hierarchical search agent skills system, comprising reusable search strategies and site-specific access procedures across sessions to augment the agents' search process.
\end{enumerate}

\section{Problem Formulation}
\label{sec:formulation}

We formulate open-domain information seeking as \textbf{relational schema completion with grounded citations}. Following relational database design, we use primary keys to identify entities, foreign keys to express cross-table relations, and normalized tables to reduce redundant or inconsistent fact representations~\citep{codd1970relational,bernstein1976synthesizing}. 

Given a natural-language request $q$, the system constructs a relational search schema
\begin{equation}
    \mathcal{S} = \bigl(\{T_m\}_{m=1}^{M},\mathcal{R}\bigr),
    \qquad
    T_m = \bigl(\mathcal{A}_m,\mathcal{P}_m\bigr),
\end{equation}
where $T_m$ defines a table with attributes $\mathcal{A}_m$ and primary-key attributes $\mathcal{P}_m$, and $\mathcal{R}$ contains foreign-key relations between tables. The task is denoted by $\mathcal{T}=(q,\mathcal{S})$.

For each table, the system must discover a set of entities $\mathcal{E}_m=\{e_{m,i}\}_{i=1}^{N_m}$ and populate a value matrix $\mathbf{Y}_m\in\mathcal{V}^{N_m\times |\mathcal{A}_m|}$. Unlike conventional table completion, every populated value must be grounded in source evidence. We therefore maintain a citation matrix $\mathbf{C}_m$ of the same shape, where each entry links a value to a source URL and an anchored excerpt:
\begin{equation}
    \mathcal{O}
    = \left\{\bigl(\mathcal{E}_m,\mathbf{Y}_m,\mathbf{C}_m\bigr)\right\}_{m=1}^{M}.
\end{equation}
The relational structure provides an explicit target for search, while the citation matrix makes each factual value independently verifiable. The completed schema is finally synthesized into a natural-language report with fine-grained inline citations.

\section{SearchOS}
\label{sec:overview}

\begin{figure}[t]
\centering
\includegraphics[width=\linewidth]{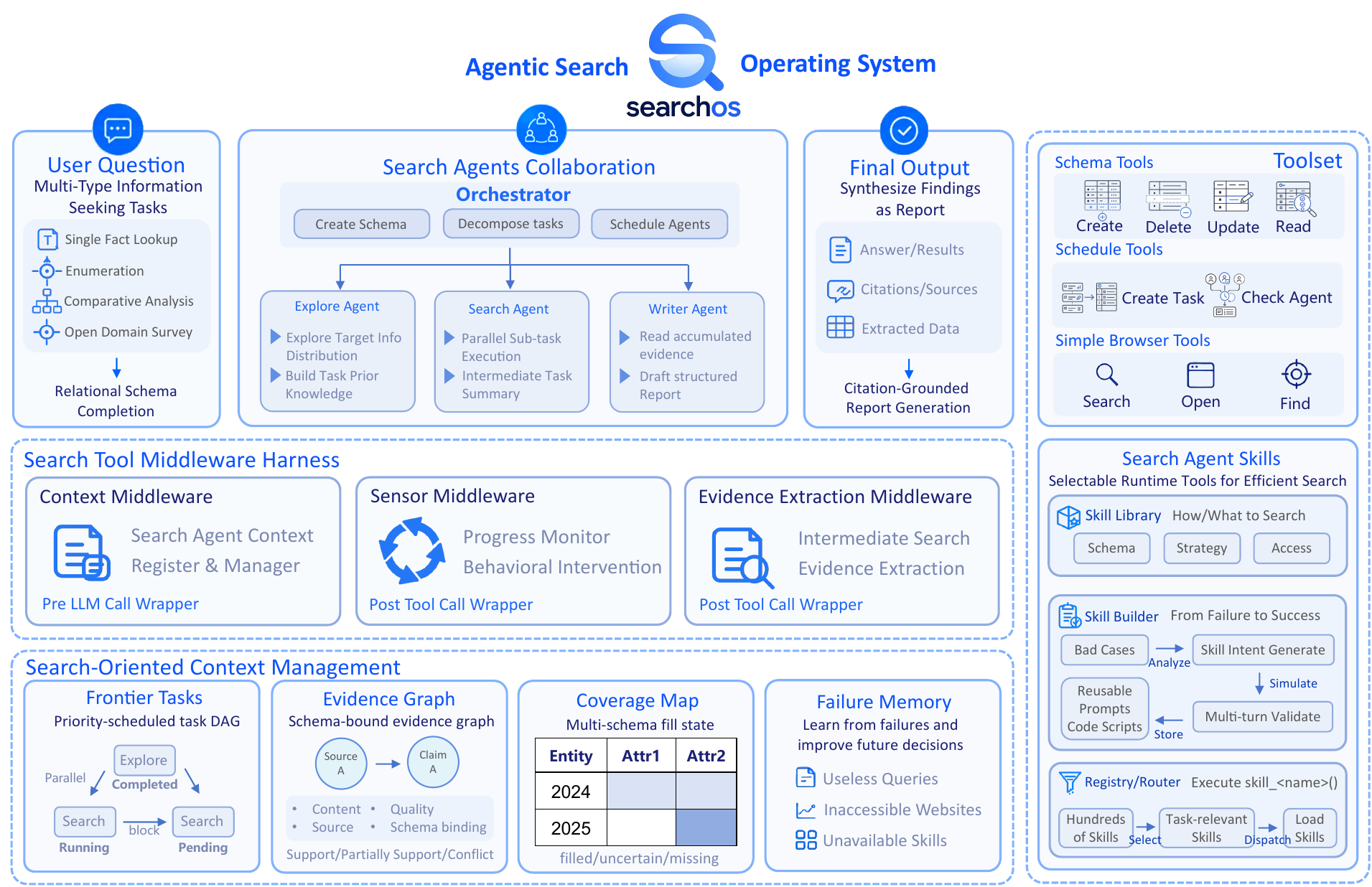}
\caption{\searchos{} architecture.}
\label{fig:overview}
\end{figure}

To keep execution aligned with the evolving search state, we implement \searchos{} as a stateful closed loop (Figure~\ref{fig:overview}). We dispatch unresolved schema gaps through the orchestrator, commit observations through SOCM, and ground each interaction through middleware. Once complete, an optional writer produces the final report.

\subsection{Search-Oriented Context Management}
\label{sec:socm}

Because long-horizon search must track unresolved schema gaps, grounded evidence, coverage, and failed attempts, we design SOCM to externalize this information as a durable shared state
\begin{equation}
    \mathcal{M}_t
    = \bigl(\mathcal{F}_t,\mathcal{G}_t,\mathcal{C}_t,\mathcal{W}_t\bigr),
\end{equation}
where $\mathcal{F}_t$ is Frontier Task, $\mathcal{G}_t$ the Evidence Graph, $\mathcal{C}_t$ the Coverage Map, and $\mathcal{W}_t$ Failure Memory. This state resides outside agent conversations and uses locked read--modify--write updates. An agent with role $r$ and assignment $z_t$ receives
\begin{equation}
    x_t^{(r)}=\phi_r(\mathcal{M}_t;z_t),
\end{equation}
rather than the full state. We use role-specific projections to expose global gaps to the orchestrator, assigned cells to search agents, and a compact summary to the writer. Each projection is regenerated from the latest state to avoid stale model context.

\subsubsection{Frontier Task}
\label{sec:frontier}
To turn schema gaps into schedulable work, we maintain Frontier Task as a dependency-aware task pool. We represent each task and the ready set as
\begin{equation}
    f_j=(\kappa_j,s_j,p_j,B_j,Z_j,a_j,n_j),\qquad
    \mathcal{R}_t=\{f_j\in\mathcal{F}_t \mid
    s_j=\textsc{Pending}\ \land\ B_j\subseteq\mathcal{T}_t\},
\end{equation}
where $\kappa_j$, $s_j$, $p_j$, $B_j$, $Z_j$, $a_j$, and $n_j$ denote type, status, priority, dependencies, target cells, assigned agent, and attempt count; $\mathcal{T}_t$ contains terminal task IDs. A task becomes pending after all dependencies terminate, then running on dispatch and completed on report collection; cycle, retry, or capacity guards may cancel it. Target-cell deduplication rejects overlapping active work, while dispatch-time revalidation closes tasks whose cells are already filled.

\subsubsection{Evidence Graph}
\label{sec:evidence}
To preserve fine-grained provenance, we store atomic findings rather than page-level summaries in the Evidence Graph:
\begin{equation}
\begin{gathered}
    \mathcal{G}_t=(\mathcal{N}_t,\mathcal{L}_t),\qquad
    g_i=(v_i,u_i,x_i,\mathbf{b}_i,\gamma_i,\tau_i,s_i), \\
    \mathcal{L}_t\subseteq\mathcal{N}_t\times\mathcal{R}_{E}\times\mathcal{N}_t,
    \qquad
    \mathcal{R}_{E}=\{\textsc{Support},\textsc{Conflict},\textsc{Refine}\}.
\end{gathered}
\end{equation}
A node $g_i\in\mathcal{N}_t$ records value $v_i$, source $u_i$, supporting span $x_i$, schema binding $\mathbf{b}_i=(m_i,e_i,a_i)$, confidence $\gamma_i$, provenance tier $\tau_i$, and status $s_i$. Rejected or superseded nodes remain for audit but do not count toward coverage. Deduplication uses binding, normalized value, and source, preserving corroborating sources. Provenance ranks anchored spans above unanchored pages and derived summaries.

\subsubsection{Coverage Map}
\label{sec:coverage}
To make progress measurable, we materialize each schema cell $c=(m,e,a)$ in the Coverage Map with status (missing, filled, uncertain, or unreachable), supporting set $H_c$, primary evidence, and a conflict flag. When several findings support a cell, we select
\begin{equation}
    g_c^*=\arg\max_{g\in H_c}
    \bigl(\tau(g),\alpha(g),\gamma(g)\bigr),
\end{equation}
in lexicographic order, where $\tau$ is provenance tier, $\alpha$ schema alignment, and $\gamma$ authority-adjusted confidence. Disagreement marks a conflict rather than overwriting evidence.

Let $\Omega_t$ be the materialized cells and $\mathcal{E}_m(t)$ the discovered rows of table $m$. Coverage is computed as
\begin{equation}
    \operatorname{Cov}(\mathcal{C}_t)=
    \frac{\sum_{c\in\Omega_t}\mathbf{1}[s(c)=\textsc{Filled}]}
    {|\Omega_t|+\sum_{m:\,|\mathcal{E}_m(t)|=0}|\mathcal{A}_m|}.
\end{equation}
The denominator keeps a declared empty table unfinished. In open-set tables, entity discovery expands $\Omega_t$, so coverage measures known rows rather than exhaustive enumeration.

\subsubsection{Failure Memory}
\label{sec:failure_memory}
Multi-agent search requires a shared record of unsuccessful actions; otherwise, one agent may repeat a path that another has already exhausted. Failure Memory stores each record as
\begin{equation}
    w_k=(\eta_k,\sigma_k,\chi_k,n_k,t_k),
    \qquad \mathcal{W}_t=\{w_k\},
\end{equation}
where $\eta_k$ is the failure type, $\sigma_k$ its task-scoped signature, $\chi_k$ the corrective guidance, $n_k$ the recurrence count, and $t_k$ the latest occurrence. Recorded patterns include uninformative or repeated queries, inaccessible sources, failed skills, dropped branches, and rejected claims. Matching failures increment $n_k$; role-specific projections expose relevant records, and recurrence strengthens ``do not retry'' guidance. Post-mortems from failed runs can seed memory for related tasks.

\subsubsection{Atomic State Updates}
SOCM updates its four memories through the same locked state interface. Evidence-producing observations jointly update the Evidence Graph and Coverage Map, while agent reports and middleware sensors update Frontier Task and Failure Memory. This coupling prevents agents from observing evidence, coverage, task status, or failure guidance from different state versions; the corresponding middleware transitions are detailed in Sections~\ref{sec:extraction_middleware} and~\ref{sec:sensors}.

\subsection{Pipeline-Parallel Multi-Agent Orchestration}
\label{sec:tools}

\paragraph{Agent roles.}
\label{sec:agent_roles}
To separate global orchestration from local search, we adopt an orchestrator--worker architecture~\citep{wu2023auto}. The \emph{orchestrator} builds the schema, prioritizes gaps, and decides when to stop; \emph{explore agents} identify candidates and sources; \emph{search agents} collect grounded evidence for scoped gaps; and the \emph{writer} produces a cited report. We coordinate these roles through SOCM rather than agent-to-agent conversation.

\paragraph{Continuous dispatch.}
To avoid idle time from synchronized batches, we apply the utilization benefits of pipeline parallelism~\citep{huang2019gpipe,narayanan2019pipedream} to role-scoped work, synchronized through SOCM. At time $t$, let $b_t$ be the number of available execution slots and $\mathcal{R}_t$ the ready set defined by Frontier Task. We dispatch
\begin{equation}
    \mathcal{D}_t
    = \operatorname{Top}_{\min(b_t,|\mathcal{R}_t|)}
      \bigl(\mathcal{R}_t;\,p\bigr),
\end{equation}
where $p$ is frontier priority. After each completion, we update SOCM, recompute $\mathcal{R}_t$, and refill the released slot. This event-driven policy overlaps roles when dependencies permit and avoids batch stragglers.

\paragraph{Toolset design.}
\label{sec:tool_system}
To keep each role's action space focused, we expose role-scoped toolsets (Table~\ref{tab:tool_overview}). We reserve schema and task mutation for the orchestrator, browsing for explore and search agents, skill loading for search agents, and outline and state access for the writer. Browser observations update $\mathcal{G}$ and $\mathcal{C}$ only through middleware.

\begin{table}[h]
\centering
\caption{Role-scoped tools. \ding{51} denotes availability.}
\label{tab:tool_overview}
\small
\begin{tabular}{l c c c c}
\toprule
\textbf{Tool Group} & \textbf{Orchestrator} & \textbf{Search} & \textbf{Explore} & \textbf{Writer} \\
\midrule
\rowcolor{gtechbg} Simple Browser (search/open/find) & -- & \ding{51} & \ding{51} & -- \\
Schema \& Entity CRUD & \ding{51} & -- & -- & -- \\
\rowcolor{gtechbg} Task Queue \& Coordination & \ding{51} & -- & -- & -- \\
Outline Management & -- & -- & -- & \ding{51} \\
\rowcolor{gtechbg} SOCM Read & -- & -- & -- & \ding{51} \\
Skill Catalog (list/load) & -- & \ding{51} & -- & \ding{51} \\
\bottomrule
\end{tabular}
\end{table}

\FloatBarrier

\paragraph{Browser interface.}
\label{sec:browser}
The browser represents search as a shared navigation stack over three operations: \texttt{search} returns result pages, \texttt{open} renders and scrolls a selected source, and \texttt{find} locates keyword-matched spans. Each stack transition yields an observation; middleware determines evidence extraction and provenance. Appendix~\ref{sec:tool_appendix} gives the complete signatures.

\subsection{Search Tool Middleware Harness}
\label{sec:middleware}

\begin{figure}[t]
\centering
\includegraphics[width=\linewidth]{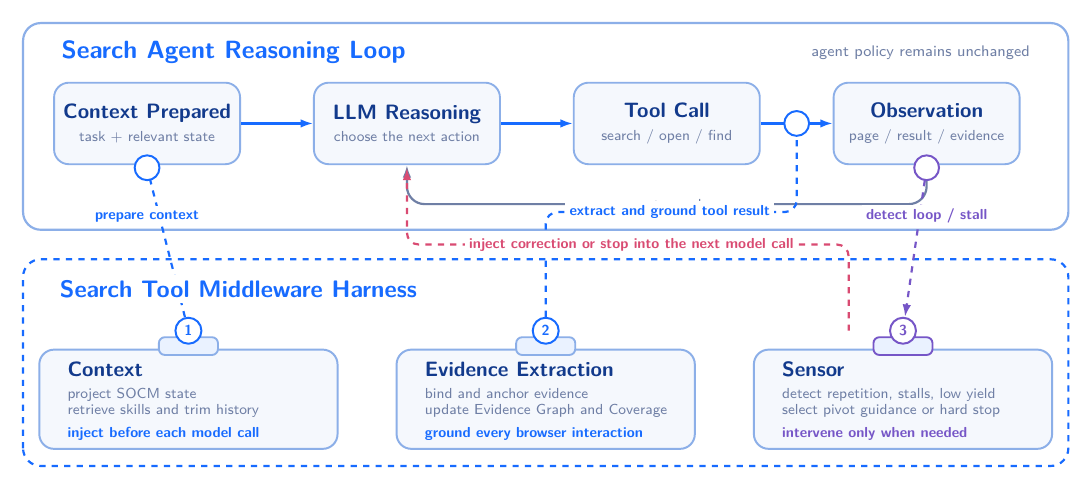}
\caption{Illustration of middleware interventions in the Search Agent loop.}
\label{fig:middleware_plugin}
\end{figure}

During long-horizon search, search agents face both model-level failures---such as losing task state or ignoring safeguards as context grows---and system-level disruptions, including tool errors, malformed outputs, stalled loops, and budget overruns. Because \textbf{prompt-level safeguards cannot reliably cover heterogeneous post-trained agents}, we introduce a system-level Search Tool Middleware Harness that intercepts the agent loop at model and tool boundaries. As illustrated in Figure~\ref{fig:middleware_plugin}, its three components form the following execution flow:
\begin{equation}
    \widetilde{h}_t^{(r)}
    =\mathcal{H}_{\mathrm{ctx}}(h_t,\mathcal{M}_t),
    \qquad
    \mathcal{M}_{t+1}
    =\mathcal{H}_{\mathrm{evidence}}(o_t,\mathcal{M}_t),
    \qquad
    a_{t+1}
    =\mathcal{H}_{\mathrm{sensor}}(\mathcal{M}_t,\mathcal{M}_{t+1},\xi_t),
\end{equation}
where $\widetilde{h}_t^{(r)}$ is the role-specific model context, $o_t$ is a tool observation, $\xi_t$ contains runtime counters, and $a_{t+1}$ is a continue, correct, or stop action. Context is prepared before inference; after tool execution, evidence is grounded and committed before sensors evaluate progress and resource use.

\subsubsection{Context Middleware}
\label{sec:harness}

Before each model call, Context Middleware composes recent history $h_t$, shared state $\mathcal{M}_t$, and retrieved skills for role $r$:
\begin{equation}
\begin{aligned}
    v_t^{(r)} &= \phi_r(\mathcal{M}_t), \qquad
    \mathcal{K}_t = \operatorname{TopK}\!\left(
        \operatorname{Retrieve}(q,\mathcal{S},v_t^{(r)})
    \right), \\
    \widetilde{h}_t^{(r)} &= h_t \oplus v_t^{(r)} \oplus \psi(\mathcal{K}_t),
\end{aligned}
\end{equation}
where $\phi_r$ projects shared state, $\mathcal{K}_t$ is the selected skill set, $\psi$ renders skills, and $\oplus$ composes context. The projection retains targets, grounded evidence, gaps, and failures without serializing the full history. Under budget pressure, older interactions are trimmed while state and active constraints remain.

\subsubsection{Evidence Extraction Middleware}
\label{sec:extraction_middleware}

Because a visited page is not necessarily evidence, we use Evidence Extraction Middleware to extract schema-bound candidates from observation $o_t$:
\begin{equation}
    \widehat{\mathcal{E}}_t
    =\operatorname{Extract}(o_t,\mathcal{S},\mathcal{C}_t),
\end{equation}
where each candidate contains an entity, attribute, value, source, and supporting span. Acceptance requires both schema binding and span anchoring:
\begin{equation}
    \mathcal{E}_t^{+}
    =\left\{z\in\widehat{\mathcal{E}}_t
      \;\middle|\;
      \operatorname{Bind}(z,\mathcal{S})=1
      \land
      \operatorname{Anchor}(z,o_t)=1
    \right\}.
\end{equation}
We commit accepted evidence in one transition:
\begin{equation}
    (\mathcal{G}_{t+1},\mathcal{C}_{t+1})
    =\mathcal{U}_{\mathrm{evidence}}
      (\mathcal{G}_t,\mathcal{C}_t,\mathcal{E}_t^{+}).
\end{equation}
This atomically records provenance and updates Coverage Map cells, making evidence capture a consequence of tool execution.

\subsubsection{Sensor Middleware}
\label{sec:sensors}

To detect stalls and enforce resource limits, we use Sensor Middleware to measure coverage and evidence progress. Let $\operatorname{Cov}(\mathcal{C}_t)$ denote grounded schema coverage:
\begin{equation}
    \Delta_t^{\mathrm{cov}}
    =\operatorname{Cov}(\mathcal{C}_t)
     -\operatorname{Cov}(\mathcal{C}_{t-1}),
    \qquad
    \Delta_t^{\mathrm{ev}}
    =|\mathcal{G}_t|-|\mathcal{G}_{t-1}|.
\end{equation}
Over window $w$, the trajectory stalls when neither quantity grows:
\begin{equation}
    s_t=\mathbb{I}\!\left[
      \sum_{j=t-w+1}^{t}\Delta_j^{\mathrm{cov}}=0
      \;\land\;
      \sum_{j=t-w+1}^{t}\Delta_j^{\mathrm{ev}}=0
    \right].
\end{equation}
Budget pressure is the maximum normalized use of iterations, searches, and time:
\begin{equation}
    \rho_t=\max\!\left(
      \frac{n_t^{\mathrm{iter}}}{B^{\mathrm{iter}}},
      \frac{n_t^{\mathrm{search}}}{B^{\mathrm{search}}},
      \frac{\tau_t}{B^{\mathrm{time}}}
    \right).
\end{equation}
Sensor Middleware computes the stall state $s_t$ and budget pressure $\rho_t$. The harness combines these signals with current gaps to decide whether to continue execution, inject a correction, request backfill, enter drain-only mode, or stop a branch. Because the harness enforces these transitions outside prompts, the same safeguards apply across models and roles.

\subsection{Hierarchical Search Skills}
\label{sec:skills}

\begin{figure}[t]
\centering
\begin{subfigure}[t]{0.49\linewidth}
    \centering
    \includegraphics[width=\linewidth]{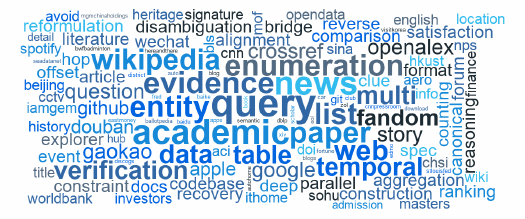}
    \caption{Frequent skill keywords.}
    \label{fig:skill_cloud}
\end{subfigure}\hfill
\begin{subfigure}[t]{0.49\linewidth}
    \centering
    \includegraphics[width=\linewidth]{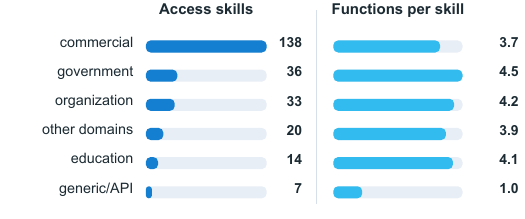}
    \caption{Access-skill counts and functions.}
    \label{fig:access_domains}
\end{subfigure}
\caption{Pre-built skills: (a) weighted keyword frequency; (b) access-skill counts and mean functions by domain.}
\label{fig:skill_library}
\end{figure}

Because search requires reusable knowledge at different operational levels, we organize it into \emph{orchestrator}, \emph{strategy}, and \emph{access} skills for global coordination, task-level methods, and source-specific retrieval. SearchOS-V1 contains 280 pre-built skills across domains (Figure~\ref{fig:skill_library}).

\paragraph{Orchestrator skills.}
Orchestrator skills are global playbooks for task decomposition, schema and row alignment, and synthesis validation; the current skill also enforces table and row-set alignment.

\paragraph{Strategy skills.}
Strategy skills encode source-independent methods for query reformulation, entity enumeration and disambiguation, structured extraction, multi-hop and temporal reasoning, aggregation, and stall recovery.

\paragraph{Access skills.}
Access skills encode source-specific retrieval and extraction for databases, government portals, encyclopedias, company sites, and media catalogs. Some combine instructions with typed executors to avoid repeated navigation and parsing; Appendix~\ref{sec:skill_example} illustrates one for Senate.gov.

\paragraph{Selection and execution.}
\label{sec:skill_hierarchy}
At startup, we inject orchestrator skills as a shared playbook. For each task, we retrieve strategy and access skills: strategies become guidance, while executable access skills become typed tools. A query-driven router prunes the access catalog before selection.


\begin{tcolorbox}[
  colback=gtechbg,
  colframe=gtechblue,
  boxrule=0.6pt,
  arc=3mm,
  left=8pt,
  right=8pt,
  top=7pt,
  bottom=7pt,
  before skip=8pt,
  after skip=0pt
]
\textbf{Scope of \searchos{}-V1.} In SearchOS-V1, we focus on externalizing search state and providing system-level infrastructure for intermediate search artifacts. In follow-up work, we will detail large-scale search agent skill synthesis from data sources, search trajectories, and user intent. \textbf{Stay tuned!} \texttwemoji{partying face}
\end{tcolorbox}

\section{Experiments}
\label{sec:exp}

\subsection{Benchmarks}

We evaluate \searchos{} on two open-domain information-seeking benchmarks as follows:

\textbf{WideSearch}~\citep{wong2025widesearch}. It targets large-scale ``wide'' information collection: it comprises 200 manually curated questions (100 English, 100 Chinese) drawn from real user queries across more than 15 domains, each requiring an agent to gather a large set of objectively verifiable atomic facts and organize them into a complete table. 

\textbf{GISA}~\citep{zhu2026gisa}. It evaluates general information-seeking assistants with 373 human-crafted queries reflecting authentic search scenarios; answers follow four structured formats (item, set, list, and table) that enable deterministic scoring, and the tasks couple deep multi-hop reasoning with broad cross-source aggregation.

\subsection{Baselines}
We compare \searchos{} against typical baselines in two categories: \emph{single-agent} methods, \textbf{ReAct}~\citep{yao2022react} and \textbf{Plan-and-Solve}~\citep{wang2023plan}, and \emph{multi-agent} systems, \textbf{A-MapReduce}~\citep{amapreduce}, \textbf{Web2BigTable}~\citep{web2bigtable}, and \textbf{Table-as-Search}~\citep{tableassearch}.

\subsection{Experimental Settings}
\textbf{Metrics.} Following the two benchmarks, we score the assembled tables against the gold standard at two granularities. \textbf{Item}-level metrics (Precision / Recall / F1) treat each answer cell independently, while \textbf{Row}-level metrics (Precision / Recall / F1) credit a row only when \emph{all} of its cells are correct. We additionally report \textbf{Exact Match (EM)}, the fraction of cases whose entire table exactly matches the gold answer. On GISA we break out F1 by question type (Table / Set / List / Item).

\textbf{Configuration.} We use \texttt{GLM-5} as the backbone for the agent roles and \texttt{Qwen3.5-35B-A3B} for Evidence Extraction. For each case we run the system three times and report the best of the three runs (Max@3), scaled by $\times100$. Unless otherwise specified, we cap each session at 50 orchestrator iterations, 8 parallel sub-agents, 20 searches per sub-agent, and a 1800-second wall-clock budget.

\subsection{Main Results}
\label{sec:main_results}

\begin{table}[t]
\centering
\caption{Main results on WideSearch and GISA. The best score in each row is in \textbf{bold}, and the second-best score is \underline{underlined}. $\Delta$ is the improvement of Ours over the strongest baseline.}
\label{tab:main}
\small
\setlength{\tabcolsep}{4.5pt}
\begin{adjustbox}{max width=\textwidth}
\begin{tabular}{ll cc ccc >{\columncolor{gtechbg}}c >{\columncolor{gtechbg}}c}
\toprule
 & & \multicolumn{2}{c}{\textbf{Single Agent}} & \multicolumn{4}{c}{\textbf{Multi-Agent System}} & \multicolumn{1}{c}{} \\
\cmidrule(lr){3-4} \cmidrule(lr){5-8}
\textbf{Benchmark} & \textbf{Metric} & ReAct & Plan-and-Solve & Table-as-Search & A-MapReduce & Web2BigTable & \textbf{Ours} & \textbf{$\Delta$} \\
\midrule
\multirow{6}{*}{WideSearch}
 & Item $\cdot$ Precision & 82.9 & \underline{83.8} & 82.4 & 83.1 & 78.3 & \textbf{83.9} & +0.1 \\
 & Item $\cdot$ Recall    & 70.2 & 72.9 & 73.5 & \underline{74.2} & 73.4 & \textbf{79.7} & +5.5 \\
 & Item $\cdot$ F1        & 72.9 & 75.2 & 75.4 & \underline{76.0} & 73.8 & \textbf{80.3} & +4.3 \\
 & Row $\cdot$ Precision  & 58.0 & \underline{58.7} & 57.1 & 56.9 & 57.5 & \textbf{59.0} & +0.3 \\
 & Row $\cdot$ Recall     & 48.8 & 50.2 & 51.6 & 49.8 & \underline{54.0} & \textbf{55.8} & +1.8 \\
 & Row $\cdot$ F1         & 50.9 & 52.2 & 52.7 & 51.4 & \underline{54.5} & \textbf{56.5} & +2.0 \\
\midrule
\multirow{5}{*}{GISA}
 & Table $\cdot$ Item $\cdot$ F1 & \underline{74.8} & 71.2 & 73.4 & 72.5 & 68.1 & \textbf{76.9} & +2.1 \\
 & Table $\cdot$ Row $\cdot$ F1  & \underline{58.1} & 50.7 & 54.1 & 52.1 & 45.3 & \textbf{59.7} & +1.6 \\
 & Set $\cdot$ F1                & 61.6 & \underline{63.1} & 60.9 & 62.5 & 56.7 & \textbf{76.5} & +13.4 \\
 & List $\cdot$ F1               & \underline{67.1} & 53.8 & 54.2 & 57.4 & 65.5 & \textbf{68.1} & +1.0 \\
 & Item $\cdot$ EM               & 0.0  & 16.7 & 16.7 & \underline{33.3} & \textbf{50.0} & \textbf{50.0} & 0.0 \\
\bottomrule
\end{tabular}
\end{adjustbox}
\end{table}

Table~\ref{tab:main} reports the main results. \searchos{} achieves the best overall performance on both benchmarks, leading on all six headline F1 metrics across the two benchmarks while remaining competitive on the precision-oriented metrics.

On \textbf{WideSearch}, \searchos{} attains the best Item-level Precision ($83.9$), Recall ($79.7$), and F1 ($80.3$), improving over the strongest baseline (A-MapReduce, $76.0$ F1) by $+4.3$ points. The gain is concentrated in recall, consistent with the design goal of \textbf{using coverage-aware dispatch to reduce missing cells}, while precision remains the highest among all methods. On the stricter Row-level metrics, where a single wrong cell invalidates the entire row, \searchos{} also leads in Precision ($59.0$), Recall ($55.8$), and F1 ($56.5$, $+2.0$ over the next-best Web2BigTable). Thus, the recall advantage extends to full-row consistency without sacrificing precision.

On \textbf{GISA}, \searchos{} leads across every question type. The gains are largest on \emph{Set} questions ($76.5$ vs.\ the best baseline's $63.1$, a $+13.4$-point improvement), where correctly enumerating a complete answer set benefits directly from the Evidence Graph and coverage-driven completeness checks. \searchos{} also tops \emph{Table} F1 at both the Item ($76.9$) and Row ($59.7$) levels and \emph{List} F1 ($68.1$), and matches the best Item-level EM ($50.0$). Overall, the results show that formulating open-domain information seeking as relational schema completion, backed by shared SOCM state, yields the most consistent gains precisely on the recall- and completeness-sensitive metrics that long-horizon search tasks demand.

\section{Ablations \& Analysis}
In this section, we present some insightful ablation and analytical experiments to further evaluate the performance of \searchos{}.
\subsection{Fixed Schemas vs. Search-Time Schema Planning}
\label{sec:schema_ablation}

We first test whether fixed schema structure suits all tasks. We first select 40 questions that can be decoupled into multi-table schema. For each question, we use \texttt{GPT-5.5} to construct semantically matched fixed single-table and multi-table schemas containing the same required attributes. Then we run \searchos{} under each fixed schema and compare them with our native autonomous schema planning with explore agent.

Table~\ref{tab:schema_ablation} reports the comparison on 40 cases. Fixed multi-table is stronger on average than fixed single-table, but wins on only 21 cases, loses on 17, and ties on two by Item F1, showing that their relative advantage is task-dependent. Even an oracle that selects the better fixed schema for each case remains below \searchos{} by $8.2$ Item F1 and $7.7$ Row F1 points.

\begin{table}[t]
\centering
\caption{Relational Schema Completion analysis. Oracle means selecting the higher-Item-F1 fixed schema per case.}
\label{tab:schema_ablation}
\begin{minipage}[c]{0.74\textwidth}
\centering
\small
\setlength{\tabcolsep}{4.5pt}
\begin{adjustbox}{max width=\linewidth}
\begin{tabular}{l ccc ccc}
\toprule
& \multicolumn{3}{c}{\textbf{Item-level}} & \multicolumn{3}{c}{\textbf{Row-level}} \\
\cmidrule(lr){2-4} \cmidrule(lr){5-7}
\textbf{Setting} & \textbf{Precision} & \textbf{Recall} & \textbf{F1} & \textbf{Precision} & \textbf{Recall} & \textbf{F1} \\
\midrule
Fixed Single-Table & 54.7 & 46.7 & 47.9 & 33.5 & 28.2 & 29.2 \\
Fixed Multi-Table & 66.9$_{\scriptscriptstyle(+12.2)}$ & 55.7$_{\scriptscriptstyle(+9.0)}$ & 58.3$_{\scriptscriptstyle(+10.4)}$ & 37.4$_{\scriptscriptstyle(+3.9)}$ & 32.6$_{\scriptscriptstyle(+4.4)}$ & 34.2$_{\scriptscriptstyle(+5.0)}$ \\
Oracle Single/Multi & \underline{70.3}$_{\scriptscriptstyle(+3.4)}$ & \underline{60.7}$_{\scriptscriptstyle(+5.0)}$ & \underline{62.4}$_{\scriptscriptstyle(+4.1)}$ & \underline{44.8}$_{\scriptscriptstyle(+7.4)}$ & \underline{39.7}$_{\scriptscriptstyle(+7.1)}$ & \underline{41.2}$_{\scriptscriptstyle(+7.0)}$ \\
\rowcolor{gtechbg}
\textbf{\searchos{}} & \textbf{76.3}$_{\scriptscriptstyle(+6.0)}$ & \textbf{68.3}$_{\scriptscriptstyle(+7.6)}$ & \textbf{70.6}$_{\scriptscriptstyle(+8.2)}$ & \textbf{52.7}$_{\scriptscriptstyle(+7.9)}$ & \textbf{47.3}$_{\scriptscriptstyle(+7.6)}$ & \textbf{48.9}$_{\scriptscriptstyle(+7.7)}$ \\
\bottomrule
\end{tabular}
\end{adjustbox}
\end{minipage}\hfill
\begin{minipage}[c]{0.23\textwidth}
\centering
{\scriptsize\textbf{SearchOS Schema Choice}}\par
\vspace{3pt}
\begin{tikzpicture}
  \draw[line width=6pt, line cap=butt, gtechblue]
    (90:0.56) arc[start angle=90,end angle=-225,radius=0.56];
  \draw[line width=6pt, line cap=butt, sparsereward]
    (-225:0.56) arc[start angle=-225,end angle=-270,radius=0.56];
  \node[align=center] at (0,0) {\scriptsize\textbf{40}\\[-2pt]\tiny tasks};
\end{tikzpicture}\par
\vspace{-2pt}
{\tiny\textcolor{gtechblue}{\rule{0.75em}{0.75em}}\ Single 87.5\%\hspace{4pt}%
\textcolor{sparsereward}{\rule{0.75em}{0.75em}}\ Multi 12.5\%}\par
\vspace{1pt}
\colorbox{gtechbg}{\parbox{0.92\linewidth}{\centering\scriptsize SearchOS dynamically selects the task-appropriate schema.}}
\end{minipage}
\end{table}

Our experiments show that \textbf{different tasks favor different table structures}. Tasks with concentrated entity attributes and simple relations may be better represented by a single table, whereas those involving one-to-many relations, multiple entity types, shared attributes, or cross-entity organization are more likely to benefit from multiple tables. Across the 40 cases, \searchos{} selected a single-table schema for 35 tasks and a multi-table schema for five. \textbf{Neither fixed structure is universally optimal.}

\textbf{The consistent advantage of \searchos{} shows that schema planning should remain part of the search process rather than be fixed in advance.} By retaining the complete search-and-planning loop, the system can organize information according to the entity types, field distributions, repetition patterns, and relations encountered during search, choosing a single-table, multi-table, or hybrid structure as appropriate and revising it when new evidence changes the information topology.

\Needspace{12\baselineskip}
\subsection{Pipeline-Parallel Scheduling Ablation}
\label{sec:pipeline-scheduling-ablation}

\begin{wraptable}{r}{0.51\textwidth}
\vspace{-0.7\baselineskip}
\centering
\caption{Round-wise scheduling results (median across 10 cases).}
\label{tab:pipeline-scheduling-rounds}
\small
\setlength{\tabcolsep}{5pt}
\renewcommand{\arraystretch}{1.05}
\begin{adjustbox}{max width=\linewidth}
\begin{tabular}{lccc}
\toprule
\textbf{Round} &
\textbf{Time $\Delta$ ($\downarrow$)} &
\textbf{Tokens $\Delta$ ($\downarrow$)} &
\textbf{Item F1 $\Delta$ (pp)} \\
\midrule
1 & $-32.6$\% & $-27.7$\% & $+2.35$ \\
2 & $-32.3$\% & $-31.2$\% & $+15.00$ \\
3 & $-28.6$\% & $-10.8$\% & $+1.85$ \\
\bottomrule
\end{tabular}
\end{adjustbox}
\vspace{-0.5\baselineskip}
\end{wraptable}

We next evaluate whether pipeline parallel continuous dispatch improves pipeline utilization over synchronized batches. We conduct three paired rounds with the same queries, model configuration, and concurrency limit ($K=8$), changing only the scheduling policy on 10 WideSearch cases. Batch scheduling waits for all tasks in a group to finish, whereas continuous scheduling immediately refills each released slot.

Table~\ref{tab:pipeline-scheduling-rounds} reports the paired results in each round. Continuous scheduling consistently reduces end-to-end time and token use while improving Item F1, showing that its advantage is stable across repeated runs.

\begin{table}[H]
\centering
\caption{Pipeline scheduling ablation on WideSearch (mean across 30 trajectories per policy).}
\label{tab:pipeline-scheduling-policies}
\small
\setlength{\tabcolsep}{7pt}
\renewcommand{\arraystretch}{1.05}
\begin{adjustbox}{max width=0.95\textwidth}
\begin{tabular}{lccccc}
\toprule
\textbf{Scheduling Policy} &
\textbf{Time (s) ($\downarrow$)} &
\textbf{Slot Utilization ($\uparrow$)} &
\textbf{Tasks/min ($\uparrow$)} &
\textbf{LLM Calls ($\downarrow$)} &
\textbf{Item F1 ($\uparrow$)} \\
\midrule
Batch (Control) & 629.13 & 34.6\% & 2.99 & 341.4 & 79.66 \\
\rowcolor{gtechbg}
\textbf{Continuous (Ours)} & \textbf{476.34} & \textbf{41.7\%} & \textbf{3.37} & \textbf{296.6} & \textbf{86.75} \\
\bottomrule
\end{tabular}
\end{adjustbox}
\end{table}

Continuous scheduling reduces average end-to-end time by $24.3\%$ and improves slot utilization and task throughput, while using fewer LLM calls and achieving higher Item F1. \textbf{These results show that continuous dispatch removes idle capacity at batch barriers and improves efficiency without sacrificing retrieval quality.}

\subsection{Middleware-Governed Search Agent Execution Analysis}
\label{sec:sensor_case_study}

Beyond aggregate scores, we further analyze whether our Search Tool Middleware Harness governance keeps long-horizon search productive after detecting stagnation.

\begin{figure}[H]
    \centering
    \includegraphics[width=0.95\textwidth]{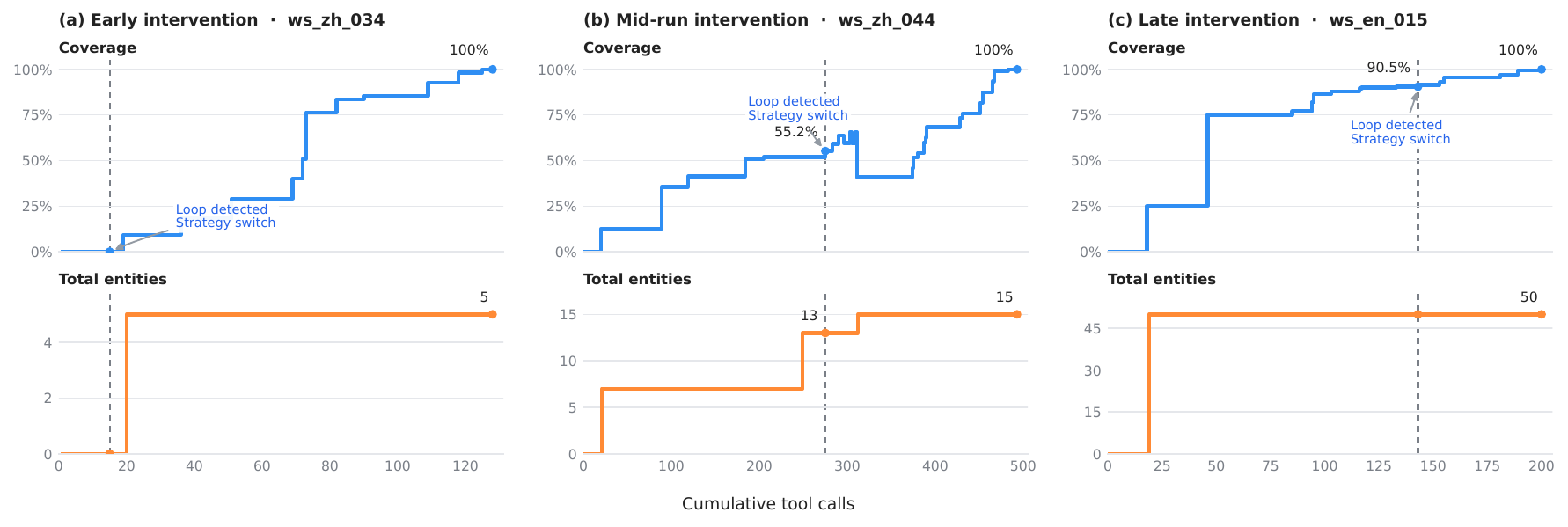}
    \caption{Middleware-governance trajectories on WideSearch.}
    \label{fig:sensor_intervention_cases}
\end{figure}

Figure~\ref{fig:sensor_intervention_cases} presents three representative WideSearch trajectories with early, mid-run, and late interventions. In each column, the upper figure shows table coverage and the lower figure shows the cumulative number of discovered entities. The dashed line marks the point at which the Loop Sensor detects a low-progress search loop and triggers a strategy switch.

Across the different cases, interventions at early, middle, and late stages all restore search progress beyond the preceding stagnation period, although they affect coverage and entity discovery to different degrees. These trajectories provide mechanism-level evidence that the Loop Sensor can redirect stalled search toward productive work.

\subsection{Skill Ablation and Efficiency}
\label{sec:skill_ablation}

We compare \searchos{} with and without hierarchical skills to assess their contribution to search quality and efficiency. Table~\ref{tab:skill_ablation} reports Max@3 over three runs per setting.

\begin{table}[H]
\centering
\caption{Skill ablation results (Max@3).}
\label{tab:skill_ablation}
\small
\setlength{\tabcolsep}{6pt}
\begin{adjustbox}{max width=\textwidth}
\begin{tabular}{l ccc ccc}
\toprule
& \multicolumn{3}{c}{\textbf{Item-level}} & \multicolumn{3}{c}{\textbf{Row-level}} \\
\cmidrule(lr){2-4} \cmidrule(lr){5-7}
\textbf{Setting} & \textbf{Precision} & \textbf{Recall} & \textbf{F1} & \textbf{Precision} & \textbf{Recall} & \textbf{F1} \\
\midrule
Without Skills & 82.2 & 78.5 & 78.3 & 56.0 & 52.4 & 53.1 \\
\rowcolor{gtechbg}
\textbf{With Skills} & \textbf{83.9} & \textbf{79.7} & \textbf{80.3} & \textbf{59.0} & \textbf{55.8} & \textbf{56.5} \\
$\Delta$ & +1.7 & +1.2 & +2.0 & +3.0 & +3.4 & +3.4 \\
\bottomrule
\end{tabular}
\end{adjustbox}
\end{table}

\paragraph{Effectiveness.}
Skills raise Item F1 by $2.0$ points and Row F1 by $3.4$ points. The larger row-level gain suggests that reusable decomposition, search, and source-access knowledge helps agents assemble coherent entity records rather than merely recover isolated cells.

\begin{center}
    \centering
    \includegraphics[width=0.95\textwidth]{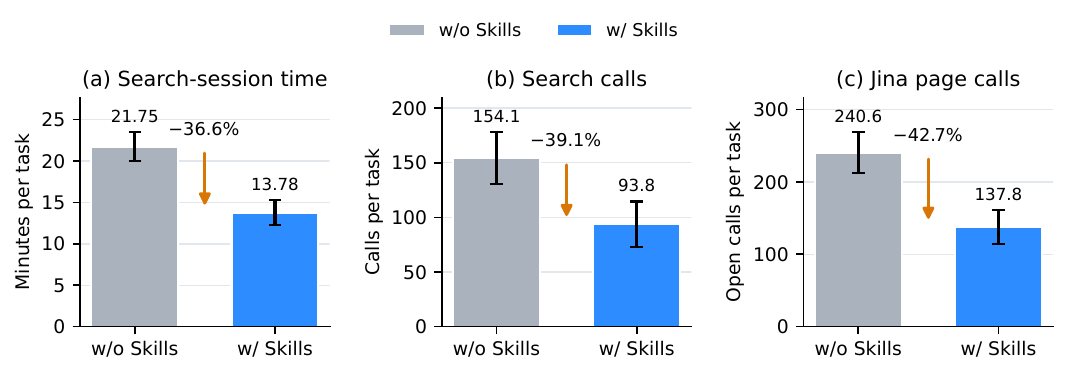}
    \captionof{figure}{Efficiency with and without skills on the same 100 WideSearch questions. Error bars show 95\% confidence intervals.}
    \label{fig:skill_efficiency}
\end{center}

\paragraph{Efficiency.}
Figure~\ref{fig:skill_efficiency} shows that skills reduce session time by $36.6\%$, search calls by $39.1\%$, and page calls by $42.7\%$. The quality gains therefore come with less exploratory trial and error rather than additional browsing. Since all skill layers are disabled together, this ablation measures their joint contribution.

\section{Related Work}
\label{sec:related}

\subsection{Agentic Information Seeking}

Tool-integrated language agents interleave search, browsing, reasoning, evidence collection, and synthesis~\citep{webgpt,yao2022react,ircot,webcpm,chen2025mindsearch,searchr1,webthinker,zheng2025deepresearcher,websailor}. Their capabilities have advanced through search policy optimization~\citep{qian2025scent,infoflow}, task decomposition~\citep{chen2025mindsearch,amapreduce}, synthetic interaction data~\citep{webshaper,websailor}, and reinforcement-based post-training~\citep{stepsearch,searchr1,gao2025turnsunlockinglonghorizonagentic,recall,zheng2025deepresearcher}. Yet progress typically remains in growing interaction histories, obscuring established evidence and unresolved questions.

Broad information seeking has also been formulated as structured or table-oriented completion~\citep{webshaper,wong2025widesearch,zhu2026gisa,tableassearch,amapreduce,web2bigtable}. Such outputs expose missing entities and attributes and support parallel collection, but do not capture task dependencies, conflicting observations, provenance, or failed attempts. \searchos{} instead formulates the task as \emph{relational schema completion with grounded citations} and maintains these intermediate artifacts outside the interaction history.

\subsection{Multi-Agent Orchestration Systems}

Multi-agent frameworks decompose complex tasks through specialized roles, message exchange, and structured workflows~\citep{wu2023auto,metagpt,agentverse,chatdev,chen2025mindsearch,magentic_one}; hierarchical approaches decouple strategic planning from specialized execution~\citep{hira}, while structured execution systems distinguish dependent from parallel sub-tasks and execute independent branches concurrently~\citep{llmcompiler,QCompiler}. However, coordination state usually remains in conversations, plans, or general task ledgers, making redundant collection, stale work, coverage gaps, and idle capacity difficult to manage explicitly during long-horizon search.

Recent advances in agent memory and context management address related challenges through reflection, memory stores, virtual context, compression, or trajectory consolidation~\citep{generative_agents,reflexion,memorybank,memgpt,jiang2023longllmlingua,agentfold}. In contrast, Search-Oriented Context Management (SOCM) externalizes intermediate search state as a Frontier Task, Evidence Graph, Coverage Map, and Failure Memory, with role-specific projections for each agent.

\subsection{Agent Harness Engineering}

Recent surveys treat the agent harness as a first-class infrastructure layer governing execution, tools, context, state, lifecycle, observability, and evaluation, while engineering reports emphasize inspectable environments, feedback loops, and context curation for reliable long-horizon agents~\citep{li2026agentharness,meng2026agentharness,lopopolo2026harness,anthropic2025context}. Prior systems instantiate parts of this layer through environment middleware, message-driven and fault-tolerant runtimes, operating-system abstractions, efficient model--tool interception, and runtime constraints~\citep{gu2024middleware,agentscope,aios,infercept,agentspec,ning2026codeharness,zhang2026web}. The \searchos{} Search Tool Middleware Harness applies these ideas to long-horizon information seeking by preparing role-specific context, validating and grounding evidence-producing observations, and monitoring repetition, stalls, and budget exhaustion.

\section{Conclusion}
\label{sec:conclusion}

In this work, we introduce \searchos{}, a multi-agent framework that externalizes long-horizon open-domain search as system-maintained state. We formulate requests as relational schema completion with grounded citations and use Search-Oriented Context Management (SOCM) to organize coverage gaps, evidence, and intermediate search artifacts. Pipeline-parallel orchestration dispatches unresolved gaps, while our Search Tool Middleware Harness prepares context, grounds observations, and handles stalls and budget limits. A hierarchical skill library supplies reusable orchestration, search-strategy, and source-access knowledge.

Experiments on WideSearch and GISA show that \searchos{} improves both completeness and efficiency. These results suggest that explicit search state, grounded evidence, and system-level execution control provide a practical foundation for reliable multi-agent information seeking. We will extend \searchos{} to broader domains and multimodal settings while improving adaptation across agents, sources, and tasks.

\newpage
\bibliographystyle{abbrvnat}
\bibliography{main}

\newpage

\begin{appendix}

\section{SOCM State Example}
\label{sec:socm_example}

This appendix provides a concrete example of the SOCM state as injected into the orchestrator's prompt during a search session.

\begin{tcolorbox}[
  title=SOCM State Snapshot (Simplified),
  colback=blue!2,
  colframe=blue!40!black,
  coltitle=white,
  colbacktitle=blue!60!black,
  fonttitle=\bfseries,
  boxrule=0.6pt,
  arc=1mm,
  breakable
]
\small
\textbf{== Frontier Memory ==}\\
Active: 3/200 | Completed: 7 | Blocked: 1\\
\begin{tabular}{llll}
\toprule
ID & Question & Kind & Status \\
\midrule
T-008 & Revenue of Company X in 2024? & search & RUNNING \\
T-009 & CEO of Company Y? & search & PENDING \\
T-010 & Compose final report & write & BLOCKED (by T-008) \\
\bottomrule
\end{tabular}

\vspace{0.5em}
\textbf{== Coverage Map ==}\\
Table: ``Fortune 500 Comparison'' | Filled: 18/30 (60\%)\\
\begin{tabular}{lccc}
\toprule
Entity & Revenue & CEO & Founded \\
\midrule
Company X & \ding{51} & \ding{51} & -- \\
Company Y & \ding{51} & -- & \ding{51} \\
Company Z & -- & -- & -- \\
\bottomrule
\end{tabular}

\vspace{0.5em}
\textbf{== Evidence Summary ==}\\
Total nodes: 24 | Active: 21 | Conflicts: 2\\
Pending resolution: Company X revenue (source A: \$1.2B vs source B: \$1.4B)
\end{tcolorbox}

\section{Executable Access Skill Case}
\label{sec:skill_example}

An executable access skill is packaged as three aligned artifacts: \texttt{skill.md} provides source-specific usage guidance, \texttt{manifest.yaml} declares the typed parameter schema exposed to the agent, and \texttt{executor.py} implements dispatch, retrieval, parsing, and normalization. The following abridged case is taken from the current Senate.gov access skill.

\begin{tcolorbox}[
  title=Access Skill Case: Senate.gov (Abridged),
  colback=blue!2,
  colframe=blue!40!black,
  coltitle=white,
  colbacktitle=blue!60!black,
  fonttitle=\bfseries,
  boxrule=0.6pt,
  arc=1mm,
  breakable
]
\footnotesize
\begin{verbatim}
# manifest.yaml
description: Fetch current U.S. Senators by state from senate.gov
params_schema:
  function:
    type: string
    required: true
    description: one of
      [get_senators_by_state, list_all_states, get_state_history]
  state_code:
    type: string
    required: false

# typed invocation selected by the agent
{"function": "get_senators_by_state", "state_code": "OK"}

# executor.py dispatch (simplified)
async def execute(params, ctx=None):
    function = params.get("function")
    async with httpx.AsyncClient(timeout=30, follow_redirects=True) as client:
        if function == "get_senators_by_state":
            return await get_senators_by_state(client, params["state_code"])
        if function == "list_all_states":
            return await list_all_states(client)
        if function == "get_state_history":
            return await get_state_history(client, params["state_code"])

# normalized output schema
{"success": true, "state_code": "OK", "state_name": "Oklahoma",
 "senators": [{"name": ..., "party": ..., "website": ...,
                "office_address": ..., "phone": ...}],
 "senator_count": 2}
\end{verbatim}
\end{tcolorbox}

The manifest constrains the agent to a small operation vocabulary, while the executor hides source-specific URL construction, HTTP error handling, and BeautifulSoup parsing. Consequently, the agent receives stable structured records even though the underlying evidence is extracted from heterogeneous Senate state-profile pages.

\section{Implementation Reference}
\label{sec:tool_appendix}

For reproducibility, this appendix records the tool interfaces summarized in Section~\ref{sec:tool_system}: the Simple Browser operations and the orchestrator's schema, task-queue, and writer tools. These interfaces are implementation choices rather than part of the core method.

\subsection{Simple Browser}

\paragraph{Browser State Abstraction.}
We model the browser state as a tuple $\mathcal{B} = (\mathcal{P}, \mathcal{H}, \sigma)$, where $\mathcal{P}$ is a page cache mapping URLs to rendered page contents, $\mathcal{H}$ is a LIFO page stack recording the navigation history, and $\sigma$ is a scroll cursor tracking the current viewport position within the top-of-stack page. Every browser operation that produces a new page---a search query, a page fetch, or a find operation---pushes the result onto $\mathcal{H}$, establishing a natural navigation history.

This stack-based design provides several advantages over a flat URL-based model:
\begin{itemize}
    \item \textbf{Contextual link resolution}: When the agent issues \texttt{open(3)}, the browser walks the stack backwards to find the most recent page containing link ID 3, resolving the numeric reference unambiguously.
    \item \textbf{Implicit back-navigation}: The stack preserves the full navigation path, enabling further trajectory analysis.
    \item \textbf{Cross-page find}: \texttt{find} results reference the \emph{source page} (the top non-find page on the stack), and subsequent \texttt{open(match\_id)} calls auto-jump to the matched line, enabling efficient in-page exploration without losing context.
\end{itemize}

\paragraph{Browser Tools.}
Table~\ref{tab:browser_tools} summarizes the three browser operations.

\begin{table}[ht]
\centering
\caption{Simple Browser tool specifications.}
\label{tab:browser_tools}
\small
\begin{tabularx}{\textwidth}{l l X}
\toprule
\textbf{Tool} & \textbf{Parameters} & \textbf{Description} \\
\midrule
\rowcolor{gtechbg} \texttt{search} & \texttt{query}: str & Web search; return URLs, titles, and snippets. \\
\texttt{open} & \texttt{id\_or\_url}: int$|$str; \texttt{loc}: int & Visit a page by URL or link ID, starting at a given line. \\
\rowcolor{gtechbg} \texttt{find} & \texttt{pattern}: str$|$list & Grep keywords from the currently open page. \\
\bottomrule
\end{tabularx}
\end{table}

\begin{figure}[ht]
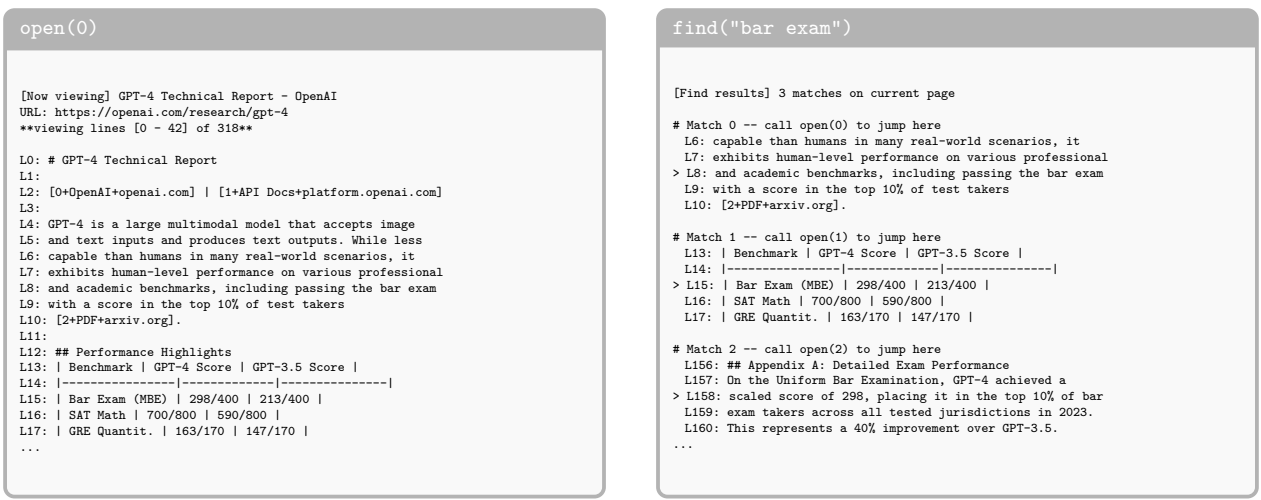

\centering
\begin{minipage}[t]{0.48\textwidth}
\begin{tcolorbox}[colback=gray!5, colframe=gray!60, title={\texttt{open(0)}}, fonttitle=\footnotesize\ttfamily, left=2pt, right=2pt, equal height group=browser]
\begin{lstlisting}[basicstyle=\ttfamily\tiny, breaklines=true, columns=fullflexible]
[Now viewing] GPT-4 Technical Report - OpenAI
URL: https://openai.com/research/gpt-4
**viewing lines [0 - 42] of 318**

L0: # GPT-4 Technical Report
L1:
L2: [0+OpenAI+openai.com] | [1+API Docs+platform.openai.com]
L3:
L4: GPT-4 is a large multimodal model that accepts image
L5: and text inputs and produces text outputs. While less
L6: capable than humans in many real-world scenarios, it
L7: exhibits human-level performance on various professional
L8: and academic benchmarks, including passing the bar exam
L9: with a score in the top 10% of test takers
L10: [2+PDF+arxiv.org].
L11:
L12: ## Performance Highlights
L13: | Benchmark      | GPT-4 Score | GPT-3.5 Score |
L14: |----------------|-------------|---------------|
L15: | Bar Exam (MBE) | 298/400     | 213/400       |
L16: | SAT Math       | 700/800     | 590/800       |
L17: | GRE Quantit.   | 163/170     | 147/170       |
...
\end{lstlisting}
\end{tcolorbox}
\end{minipage}%
\hfill
\begin{minipage}[t]{0.48\textwidth}
\begin{tcolorbox}[colback=gray!5, colframe=gray!60, title={\texttt{find("bar exam")}}, fonttitle=\footnotesize\ttfamily, left=2pt, right=2pt, equal height group=browser]
\begin{lstlisting}[basicstyle=\ttfamily\tiny, breaklines=true, columns=fullflexible]
[Find results] 3 matches on current page

# Match 0 -- call open(0) to jump here
  L6:  capable than humans in many real-world scenarios, it
  L7:  exhibits human-level performance on various professional
> L8:  and academic benchmarks, including passing the bar exam
  L9:  with a score in the top 10% of test takers
  L10: [2+PDF+arxiv.org].

# Match 1 -- call open(1) to jump here
  L13: | Benchmark      | GPT-4 Score | GPT-3.5 Score |
  L14: |----------------|-------------|---------------|
> L15: | Bar Exam (MBE) | 298/400     | 213/400       |
  L16: | SAT Math       | 700/800     | 590/800       |
  L17: | GRE Quantit.   | 163/170     | 147/170       |

# Match 2 -- call open(2) to jump here
  L156: ## Appendix A: Detailed Exam Performance
  L157: On the Uniform Bar Examination, GPT-4 achieved a
> L158: scaled score of 298, placing it in the top 10% of bar
  L159: exam takers across all tested jurisdictions in 2023.
  L160: This represents a 40% improvement over GPT-3.5.
...
\end{lstlisting}
\end{tcolorbox}
\end{minipage}
\caption{Browser tool outputs as seen by a search agent. \textbf{Left}: \texttt{open()} renders a page with line-numbered markdown and bracket-style link references (\texttt{[id\textdagger{}title\textdagger{}domain]}). \textbf{Right}: \texttt{find()} returns numbered matches with context; calling \texttt{open(match\_id)} jumps to the source line.}
\label{fig:browser_example}
\end{figure}

Figure~\ref{fig:browser_example} illustrates the rendered outputs an agent receives from \texttt{open()} and \texttt{find()} calls.

\subsection{Schema and Entity Management Tools}
\label{sec:schema_tools}

The orchestrator manages the relational search schema $\mathcal{S}$ through a suite of CRUD tools that operate on the Coverage Map (Section~\ref{sec:coverage}). Table~\ref{tab:schema_tools} lists the full tool set.

\begin{table}[ht]
\centering
\caption{Schema and entity management tools (orchestrator only).}
\label{tab:schema_tools}
\small
\begin{tabularx}{\textwidth}{l l X}
\toprule
\textbf{Tool} & \textbf{Key Parameters} & \textbf{Description} \\
\midrule
\rowcolor{gtechbg} \texttt{create\_schema} & \texttt{tables}, \texttt{relations} & Define schema with tables, attributes, PKs, and FKs. \\
\texttt{add\_entities} & \texttt{entities}, \texttt{attributes} & Add rows with optional inline values (stored as evidence). \\
\rowcolor{gtechbg} \texttt{remove\_entities} & \texttt{entities}, \texttt{reason} & Soft-delete rows with mandatory justification. \\
\texttt{edit\_entities} & \texttt{edits} & Rename PKs or edit cells. Preserves evidence refs. \\
\rowcolor{gtechbg} \texttt{inspect\_table} & \texttt{table\_id}, \texttt{status} & Return row list by status (filled/partial/empty). \\
\bottomrule
\end{tabularx}
\end{table}

\subsection{Task Queue and Agent Coordination Tools}
\label{sec:task_tools}

The orchestrator manages the Frontier Memory (Section~\ref{sec:frontier}) and sub-agent lifecycle through the tools listed in Table~\ref{tab:task_tools}.

\begin{table}[ht]
\centering
\caption{Task queue and agent coordination tools (orchestrator only).}
\label{tab:task_tools}
\small
\begin{tabularx}{\textwidth}{l l X}
\toprule
\textbf{Tool} & \textbf{Key Parameters} & \textbf{Description} \\
\midrule
\rowcolor{gtechbg} \texttt{enqueue\_tasks} & \texttt{items} & Append tasks with priority, target table, and dependencies. \\
\texttt{check\_agents} & \texttt{timeout}: float & Block until $\geq$1 agent completes; return reports. \\
\rowcolor{gtechbg} \texttt{stop\_task} & \texttt{task\_ids}, \texttt{reason} & Cancel tasks and their running sub-agents. \\
\bottomrule
\end{tabularx}
\end{table}

\Needspace{14\baselineskip}
\subsection{Writer Tools}
\label{sec:writer_tools}

The writer agent operates on a structured outline rather than producing free-form text, ensuring consistent organization and complete citation coverage. Table~\ref{tab:writer_tools} lists the writer-specific tools.

\begin{table}[H]
\centering
\caption{Writer agent tools for outline and section management.}
\label{tab:writer_tools}
\small
\begin{tabularx}{\textwidth}{l l X}
\toprule
\textbf{Tool} & \textbf{Key Parameters} & \textbf{Description} \\
\midrule
\rowcolor{gtechbg} \texttt{read\_outline} & \texttt{section\_id}, \texttt{toc\_only} & Return TOC or full section content. \\
\texttt{update\_outline} & \texttt{ops} (JSON array) & Batch add/remove/reorder sections. \\
\rowcolor{gtechbg} \texttt{write\_section} & \texttt{section\_id}, \texttt{cited\_evidence\_ids} & Write section with mandatory citations. \\
\texttt{edit\_section} & \texttt{section\_id}, \texttt{old}, \texttt{new} & Exact substring replacement for surgical edits. \\
\rowcolor{gtechbg} \texttt{annotate\_section} & \texttt{section\_id}, \texttt{annotation} & Attach annotation to a section. \\
\bottomrule
\end{tabularx}
\end{table}

In addition, the writer agent has read-only access to the full SOCM state and the skill catalog, enabling it to ground its writing in the latest evidence.

\Needspace{12\baselineskip}
\section{Agent Trajectory Case Studies}
\label{sec:trajectory_cases}

This section presents process-level case studies from complete WideSearch trajectories. We summarize the logged decision rationale, tool action, and observable state transition at each consequential step rather than reproducing unabridged token-level reasoning. The cases were selected to expose different behaviors---parallel enrichment, scope auditing, and recovery from inaccessible sources---and are illustrative rather than aggregate evidence.

\subsection{Detailed Case: Spotify 2024 Rankings}
\label{sec:case_spotify_trajectory}

\textbf{Task and outcome.} Case \texttt{ws\_en\_022} asks for the global and U.S. top-ten songs on Spotify in 2024, enriched with artist, language, songwriter, producer, and release date. The run finishes in 362 seconds with 221 evidence nodes, 100\% known-cell coverage, 97.5 Item F1, and 85.0 Row F1.

\paragraph{Decision trace.}
Table~\ref{tab:spotify_decision_trace} reconstructs the consequential intermediate decisions from the stored trajectory. The key design choice is to establish the two authoritative ranked lists before dispatching metadata enrichment, so each sub-agent writes into a stable composite key rather than independently rediscovering row identities.

\begin{table}[ht]
\centering
\caption{Condensed agent trajectory for \texttt{ws\_en\_022}. Rationale is summarized from logged steps; state values are taken from SOCM deltas.}
\label{tab:spotify_decision_trace}
\footnotesize
\setlength{\tabcolsep}{4pt}
\begin{tabularx}{\textwidth}{c p{0.25\textwidth} p{0.24\textwidth} X}
\toprule
\textbf{Stage} & \textbf{Decision rationale} & \textbf{Action} & \textbf{Observable state} \\
\midrule
1. Explore & First identify an authoritative source that fixes both top-ten lists. & Dispatch one Explore Agent to inspect Spotify Wrapped and chart sources. & No schema; coverage 0\%. \\
\rowcolor{gtechbg}
2. Structure & The official lists expose two row dimensions: chart category and rank. & Create one $20\times8$ table with primary key \texttt{[Category, Rank]}; seed both lists. & 40/160 cells known; coverage 25\%. \\
3. Parallelize & Metadata lookup is independent after row identities are fixed. & Split the 20 songs into four five-song tasks; dispatch four Search Agents with the Spotify access skill. & Four agents run concurrently. \\
\rowcolor{gtechbg}
4. Monitor & Merge completed shards without waiting for the slowest agent. & Repeatedly call \texttt{check\_agents}; retain shared Coverage Map updates. & Coverage 25.0$\rightarrow$68.8$\rightarrow$89.4$\rightarrow$100.0\%. \\
5. Audit & Full known-cell coverage does not guarantee formatting consistency or scope completion. & Inspect all 20 rows; compare known row set to the two official top-ten lists. & 20 expected rows present; zero missing rows. \\
\rowcolor{gtechbg}
6. Repair & Two release dates use inconsistent separators. & Apply two targeted \texttt{edit\_entities} operations, then export. & 20 complete rows; 221 evidence nodes. \\
\bottomrule
\end{tabularx}
\end{table}

\paragraph{SOCM evolution.}
The orchestrator receives compact state snapshots while the Search Agents work. Early in the run, all 20 seeded rows have gaps. A mid-run snapshot reports only six rows with missing or uncertain columns. The final snapshot reports zero gaps, while explicitly warning that ``all known rows filled'' is not proof that the requested row set is exhaustive. This warning triggers the Stage-5 row-set audit rather than immediate termination.

\begin{tcolorbox}[
  title=Coverage and Governance Events (Condensed),
  colback=blue!2,
  colframe=blue!40!black,
  coltitle=white,
  colbacktitle=blue!60!black,
  fonttitle=\bfseries,
  boxrule=0.6pt,
  arc=1mm
]
\small
\begin{tabularx}{\textwidth}{l X r}
\textbf{Moment} & \textbf{SOCM / middleware signal} & \textbf{Coverage} \\
After schema creation & 20 known rows; all 20 have missing attributes. & 25.0\% \\
First shard completes & Five rows receive title, singer, language, credits, and date evidence. & 68.8\% \\
Mid-run snapshot & 20 known rows; six still have missing or uncertain attributes. & 89.4\% \\
Search stalls & Loop Sensor emits four \texttt{bare\_search\_without\_progress} reminders across two agents. & -- \\
Final snapshot & 20 known rows; zero gaps; row-set audit still required. & 100.0\% \\
\end{tabularx}
\end{tcolorbox}

\paragraph{Output excerpt.}
The same song may occur in both charts, but the composite key preserves its category-specific rank while shared metadata is independently grounded.

\begin{table}[ht]
\centering
\caption{Selected final rows from \texttt{ws\_en\_022}.}
\label{tab:spotify_output_excerpt}
\small
\begin{tabular}{c c l l l}
\toprule
\textbf{List} & \textbf{Rank} & \textbf{Title} & \textbf{Singer} & \textbf{Release Date} \\
\midrule
Global & 1 & Espresso & Sabrina Carpenter & 2024/04/11 \\
Global & 2 & Beautiful Things & Benson Boone & 2024/01/19 \\
U.S. & 1 & Espresso & Sabrina Carpenter & 2024/04/11 \\
U.S. & 2 & Not Like Us & Kendrick Lamar & 2024/05/04 \\
\bottomrule
\end{tabular}
\end{table}

\subsection{Additional Process Cases}

\paragraph{False saturation followed by scope repair (\texttt{ws\_en\_016}).}
The task enumerates Michael Phelps's individual-event medals across the Olympic Games and World Aquatics Championships. The orchestrator initially splits the work by competition family. After both agents return, known-cell coverage reaches 100\%, but the table contains only 16 rows while the agent reports imply roughly 31 eligible events. Instead of treating saturation as completion, the orchestrator inspects the row set, identifies missing years and competitions, and dispatches targeted backfill tasks. The table grows to 33 and then 35 rows. A final verification pass detects a duplicate 2003 World Championships 400m individual-medley record, removes the incorrect \texttt{4:11.04} row, adds the missing 2005 100m butterfly result, and corrects medalists in several 200m freestyle rows. This trajectory illustrates why Coverage Map saturation is paired with explicit scope auditing.

\begin{tcolorbox}[
  title=Key Logged Decisions for \texttt{ws\_en\_016},
  colback=gray!4,
  colframe=gray!65,
  fonttitle=\bfseries,
  boxrule=0.5pt,
  arc=1mm,
  breakable
]
\footnotesize
\textbf{Observation:} coverage is 100\%, but only 16 rows are present; two completed agents report approximately 31 eligible medal events.\\
\textbf{Decision:} audit row scope before synthesis.\\
\textbf{Action:} inspect the table, then dispatch Olympic and World-Aquatics backfill agents.\\[0.3em]
\textbf{Observation:} two rows claim the 2003 400m individual medley with times \texttt{4:09.09} and \texttt{4:11.04}.\\
\textbf{Decision:} treat the duplicate identity as a verification target.\\
\textbf{Action:} verify against championship sources and remove the incorrect row.\\[0.3em]
\textbf{Outcome:} 35 final rows, 486 evidence nodes, 80.1 Item F1.
\end{tcolorbox}

\paragraph{Blocked publisher pages and bounded recovery (\texttt{ws\_en\_023}).}
This task enumerates migration-related articles from three journals over 2020--2024. The orchestrator first corrects an ambiguous journal name, then assigns one Search Agent per journal. After the initial pass, it removes an out-of-scope 2026 article and dispatches targeted metadata tasks for five incomplete rows. Wiley and JSTOR pages repeatedly return access checks or HTTP 402 responses. The affected agents pivot to Crossref, OpenAlex, and academic-paper skills; the Loop Sensor records two explicit strategy switches for one stalled agent. When the remaining sources and per-agent budgets are exhausted, the system preserves the unresolved cells instead of fabricating metadata. Despite these access failures, the run returns 57 rows with 96.3\% coverage and 93.8 Item and Row F1.

\begin{table}[ht]
\centering
\caption{Process summary for the three trajectory cases. Scores are scaled by $\times100$.}
\label{tab:trajectory_case_summary}
\small
\begin{adjustbox}{max width=\textwidth}
\begin{tabular}{l l c c c c}
\toprule
\textbf{Case} & \textbf{Primary behavior} & \textbf{Tasks} & \textbf{Evidence} & \textbf{Item F1} & \textbf{Row F1} \\
\midrule
\texttt{ws\_en\_022} & Parallel enrichment + final audit & 5 & 221 & 97.5 & 85.0 \\
\rowcolor{gtechbg}
\texttt{ws\_en\_016} & False-saturation repair + verification & 8 & 486 & 80.1 & 28.2 \\
\texttt{ws\_en\_023} & Source fallback + bounded completion & 11 & 302 & 93.8 & 93.8 \\
\bottomrule
\end{tabular}
\end{adjustbox}
\end{table}

\paragraph{Cross-case observations.}
The three traces expose complementary aspects of the system that are hidden by final-answer scores:
\begin{itemize}
    \item \textbf{Explore before committing to row identities.} In the Spotify case, an authoritative list establishes the complete key space before metadata agents are launched. This prevents independently discovered song records from drifting across the two chart categories.
    \item \textbf{Cell saturation and scope completion are separate conditions.} In the Phelps case, 100\% coverage over 16 known rows is correctly rejected as completion because the expected event count is much larger. Coverage controls attribute filling; an explicit row-set audit controls recall.
    \item \textbf{Recovery remains bounded and evidence preserving.} In the journal case, sensors encourage source and tool changes after repeated failures, while resource limits eventually stop unproductive search. Unresolved cells remain visible instead of being filled from unsupported inference.
\end{itemize}

Together, these cases show the intended control loop: exploration defines a defensible search space, parallel agents accumulate evidence into shared structured state, SOCM exposes concrete residual gaps, and governance redirects or stops work according to observable progress.

\clearpage
\section{SearchOS Case Walkthroughs}
\label{sec:interface_case_walkthroughs}

The following interface snapshots are ordered by case number and then by step number. Each row contains two snapshots.

\newcommand{\casepanel}[2]{%
\begin{minipage}[t]{0.485\textwidth}
  \centering
  \includegraphics[width=\linewidth]{#1}\\[-0.15em]
  {\scriptsize\textbf{#2}}
\end{minipage}%
}

\subsection{Case 1: Beijing Travel Planning}

\noindent\casepanel{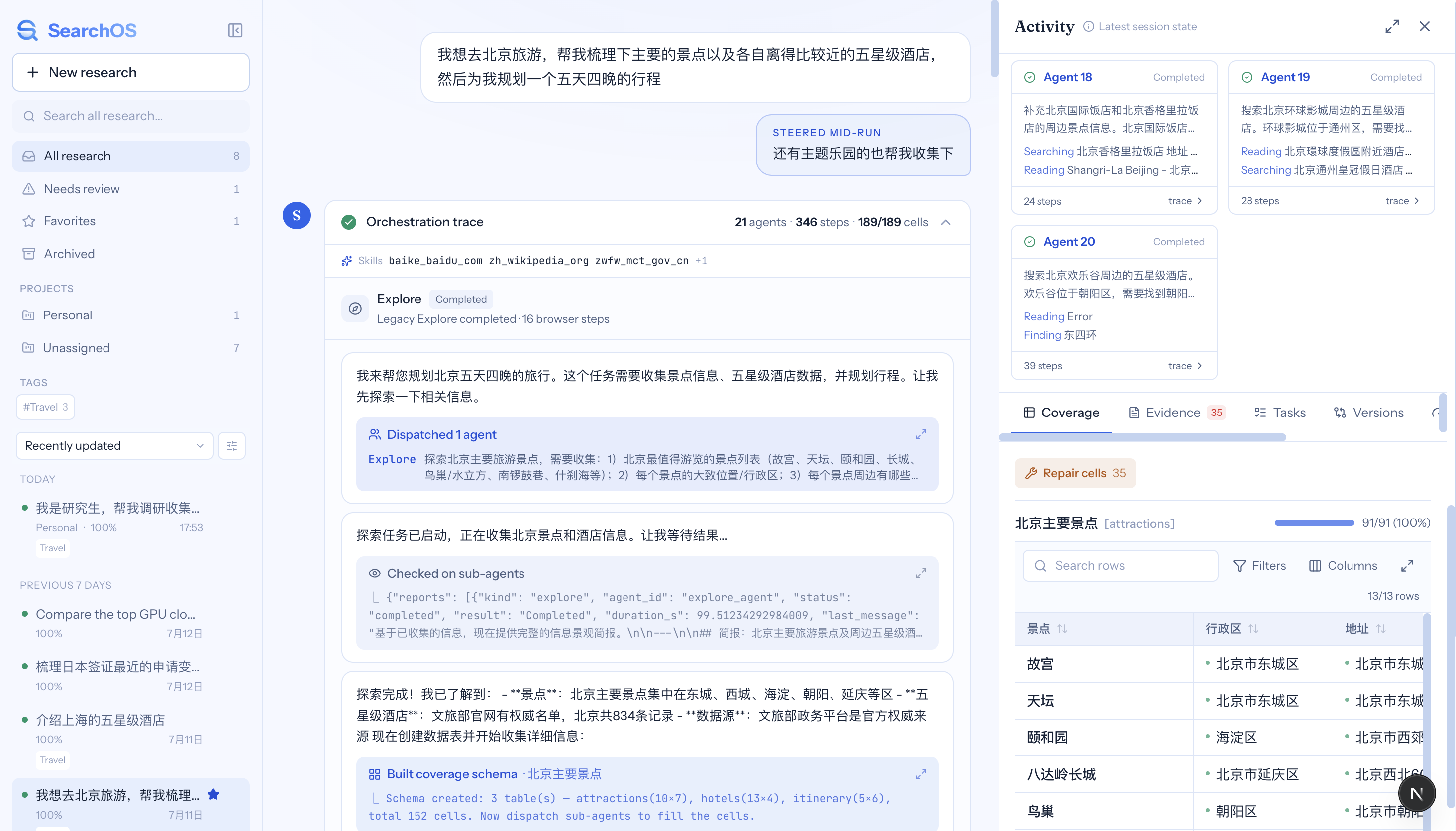}{Case 1-0}\hfill
\casepanel{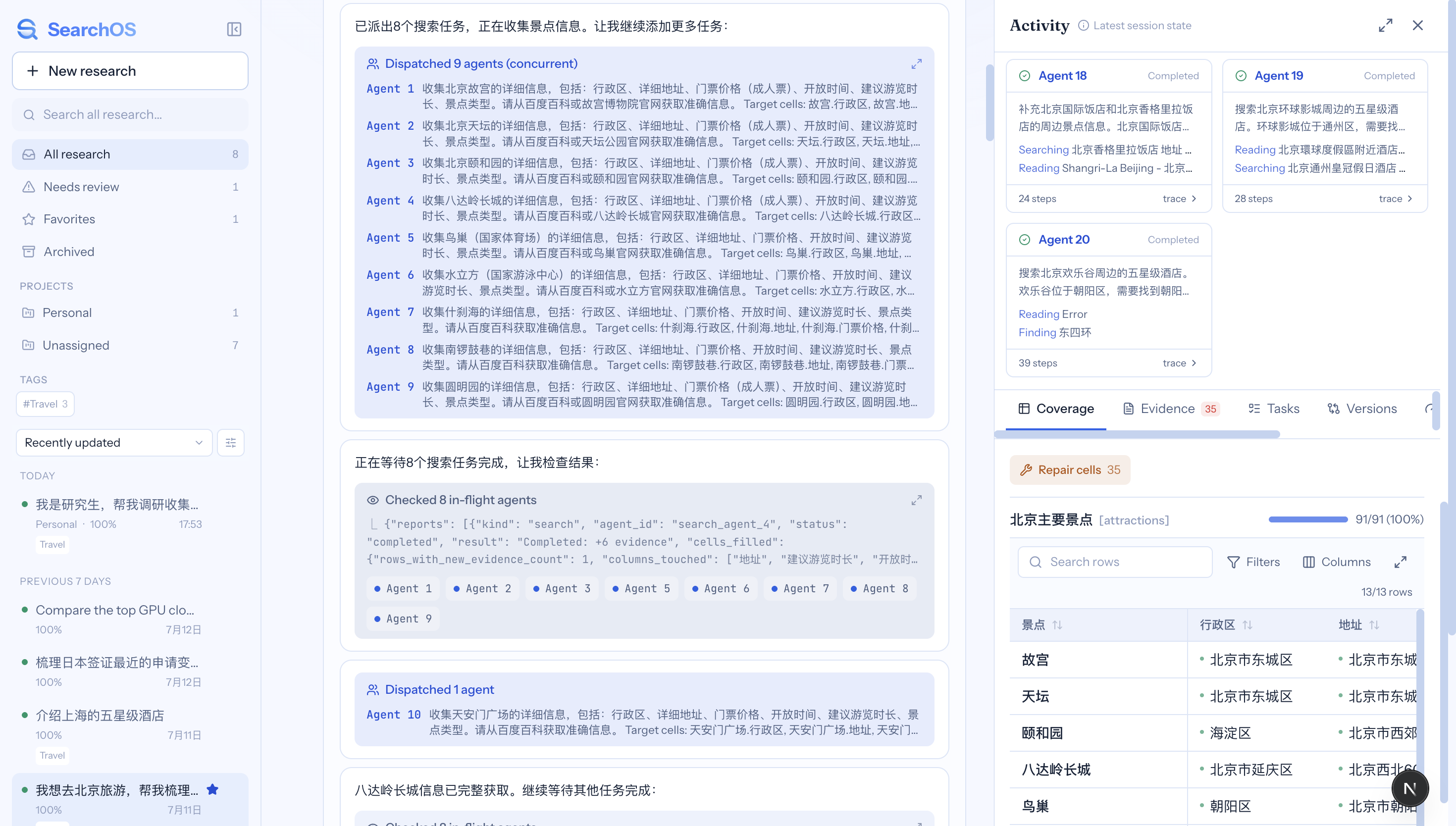}{Case 1-1}

\vspace{0.6em}
\noindent\casepanel{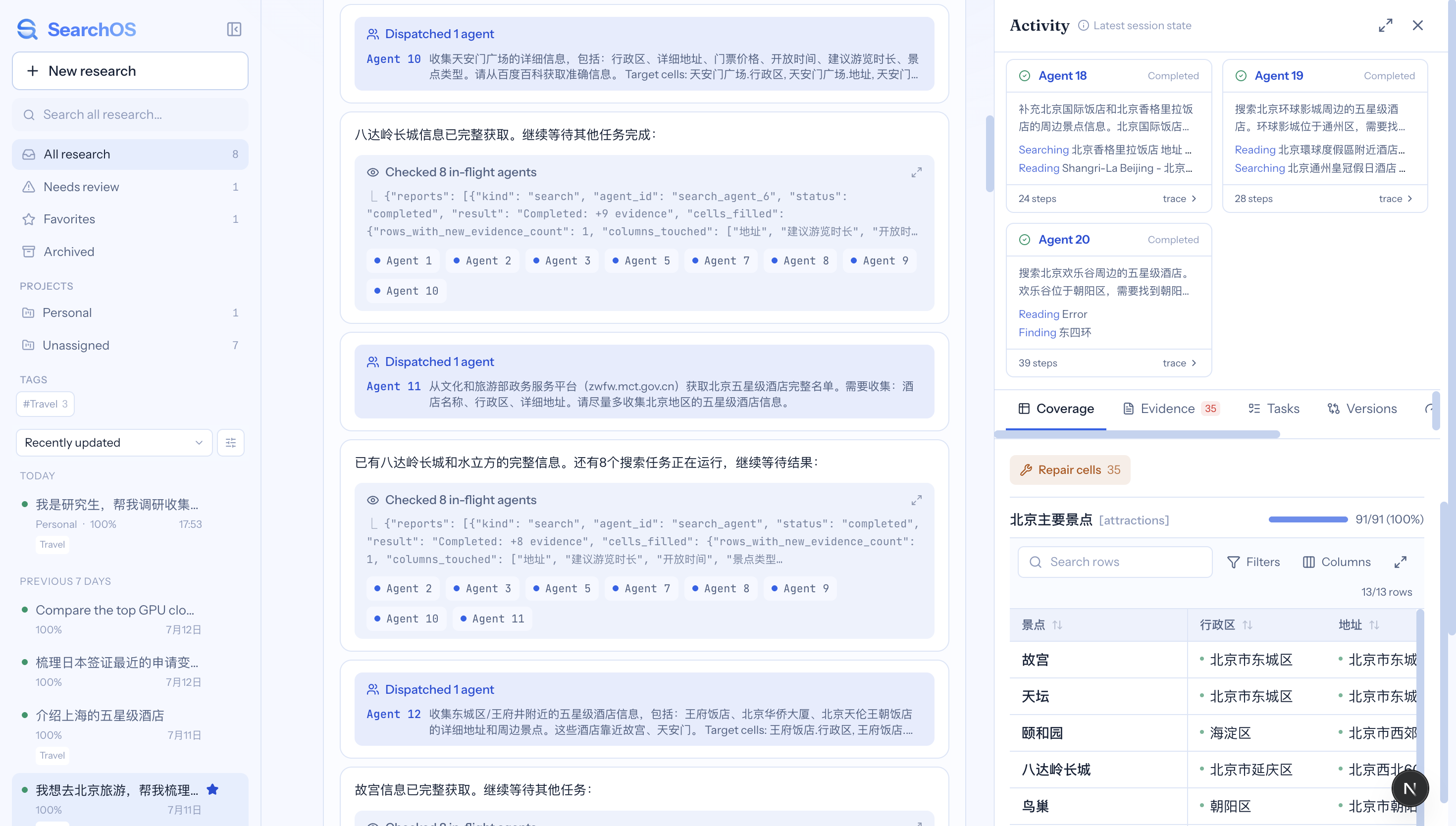}{Case 1-2}\hfill
\casepanel{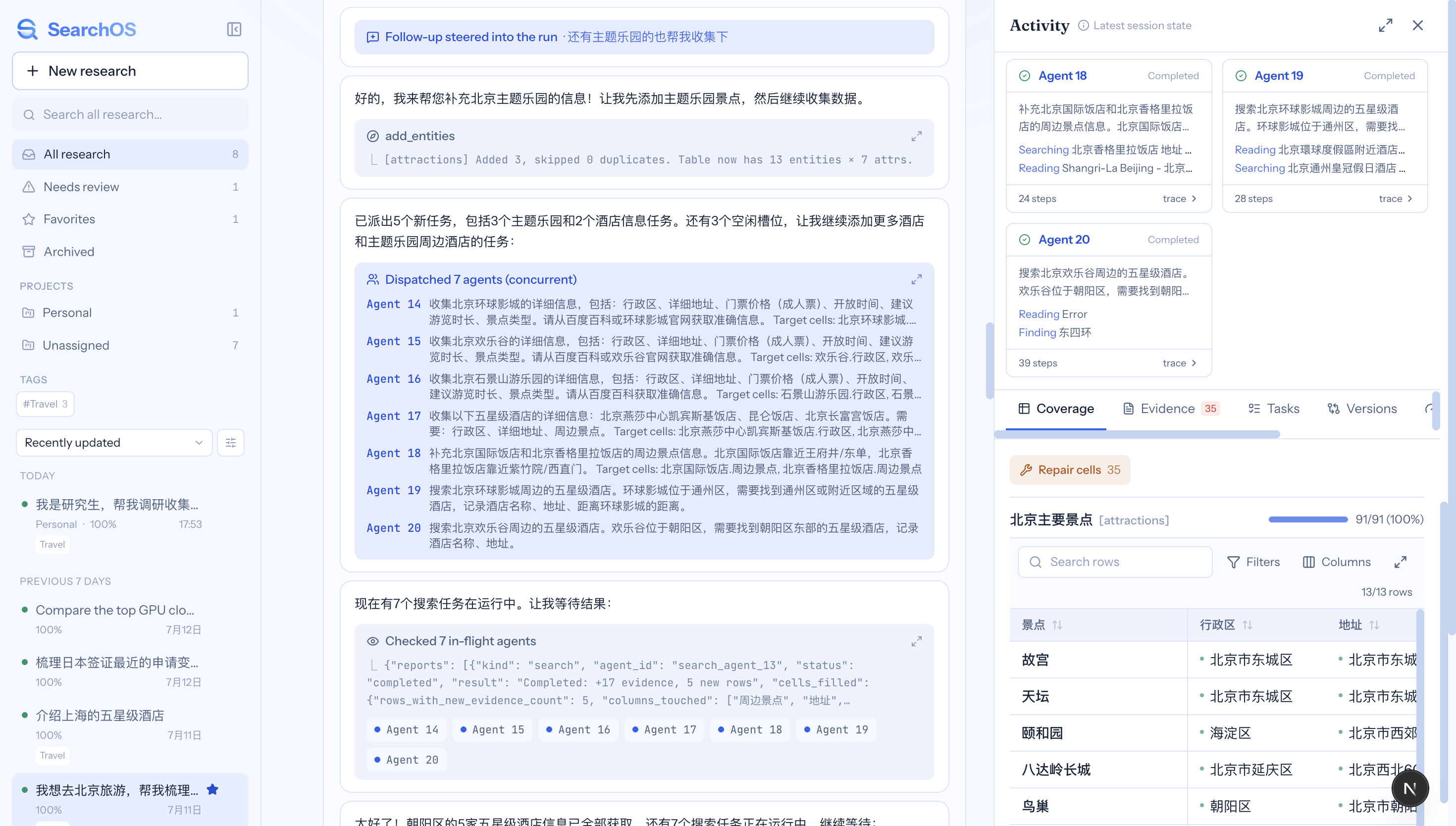}{Case 1-3}

\vspace{0.6em}
\noindent\casepanel{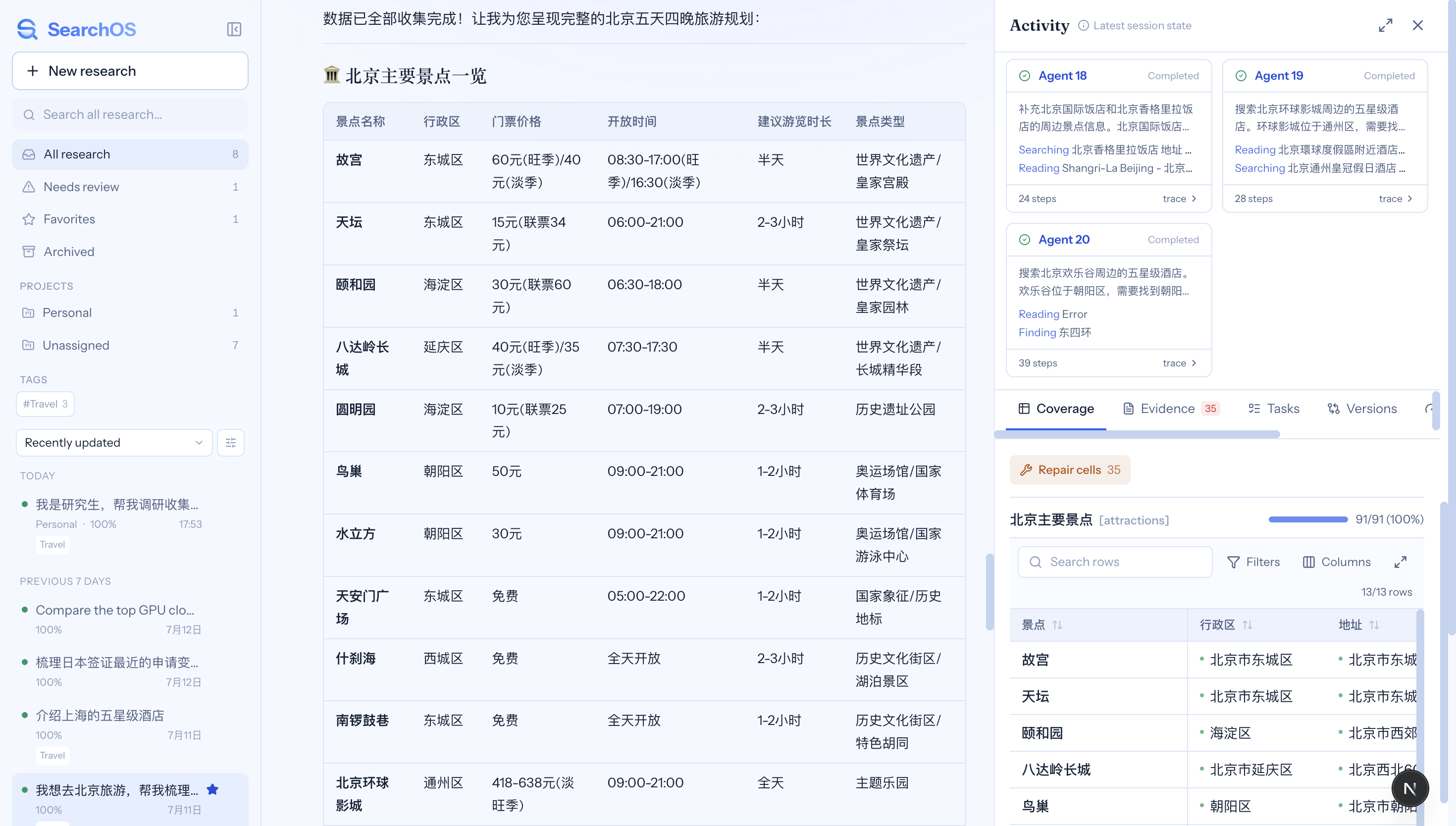}{Case 1-4}\hfill
\casepanel{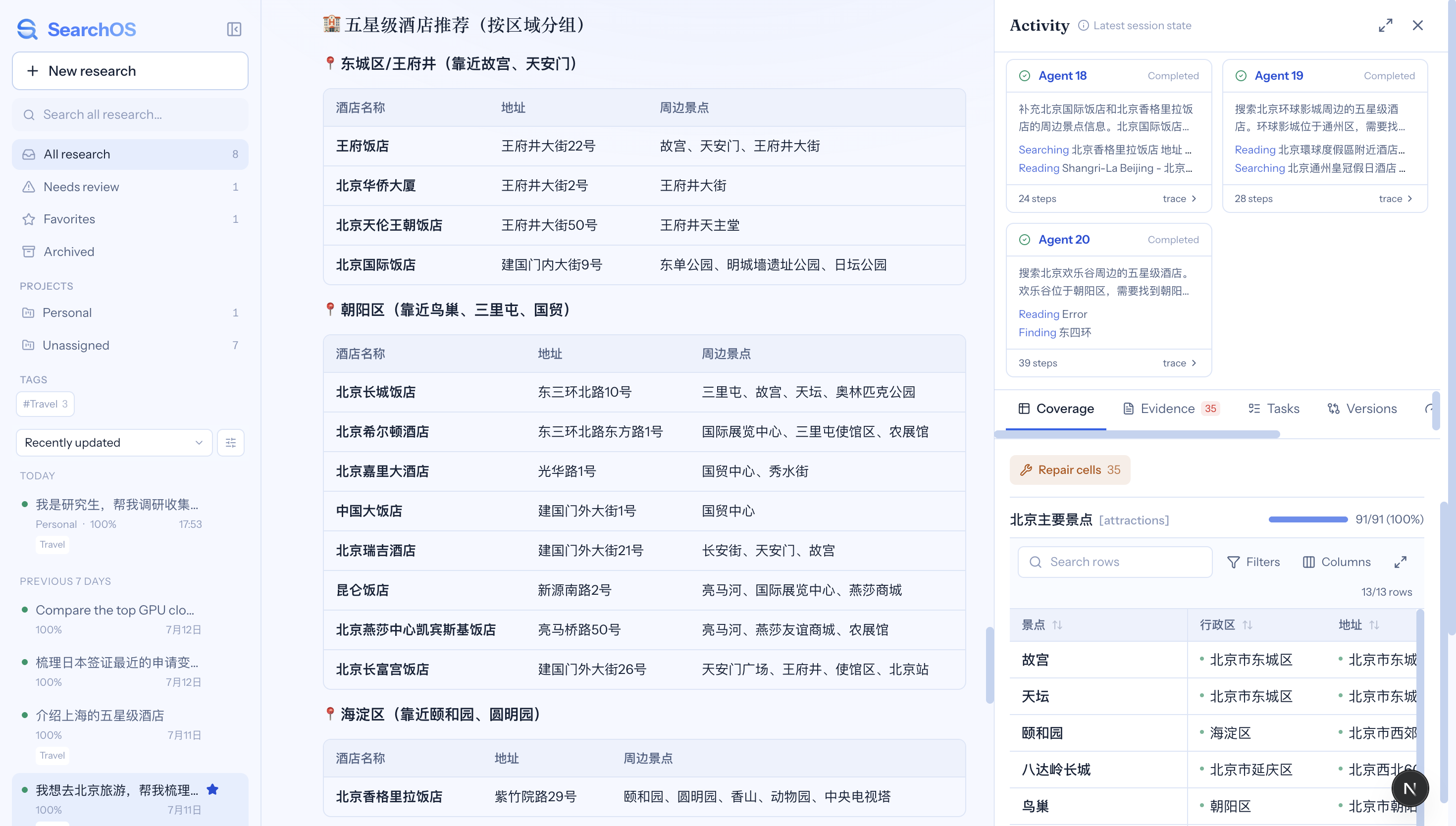}{Case 1-5}

\clearpage
\noindent\casepanel{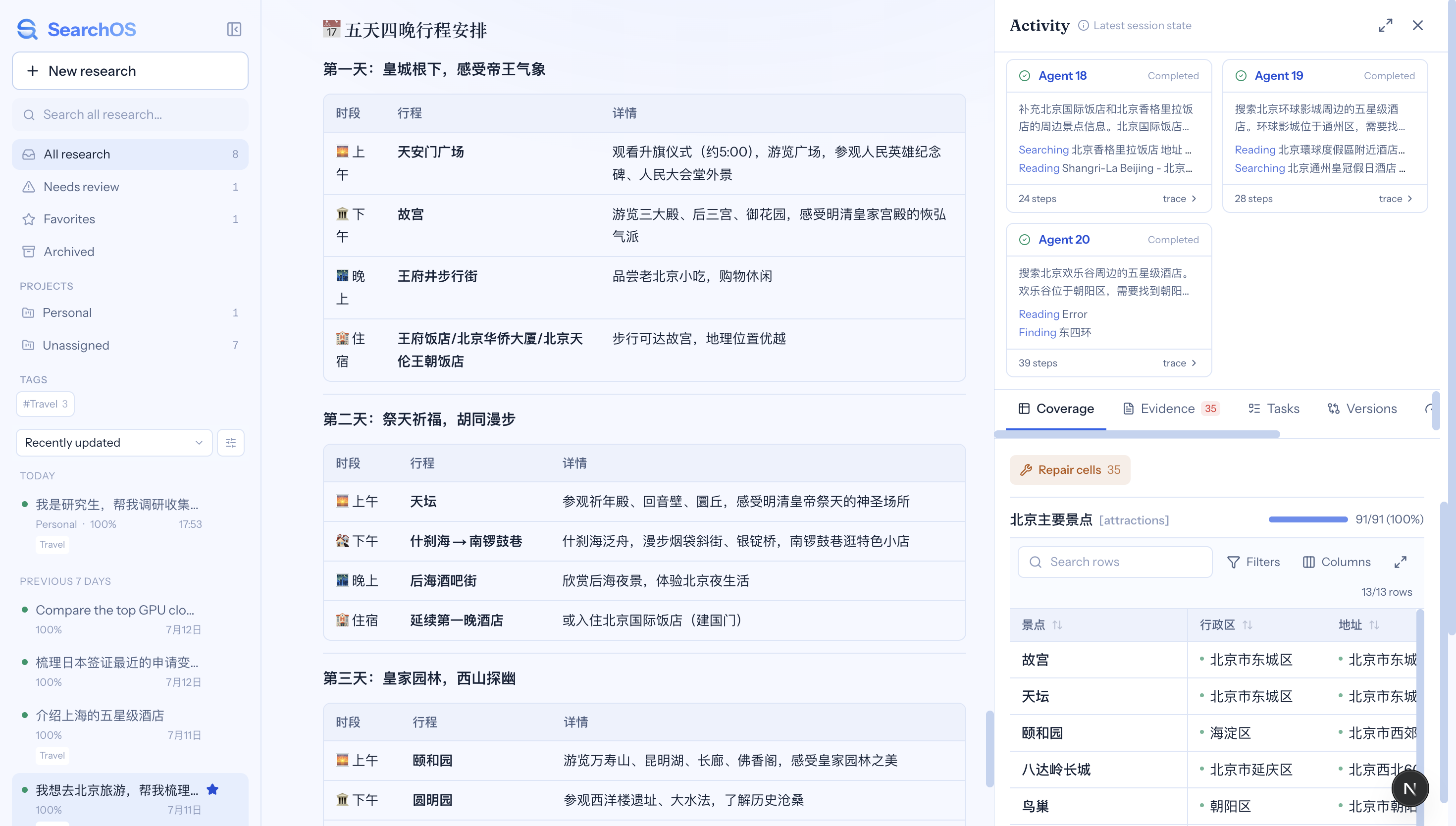}{Case 1-6}\hfill
\casepanel{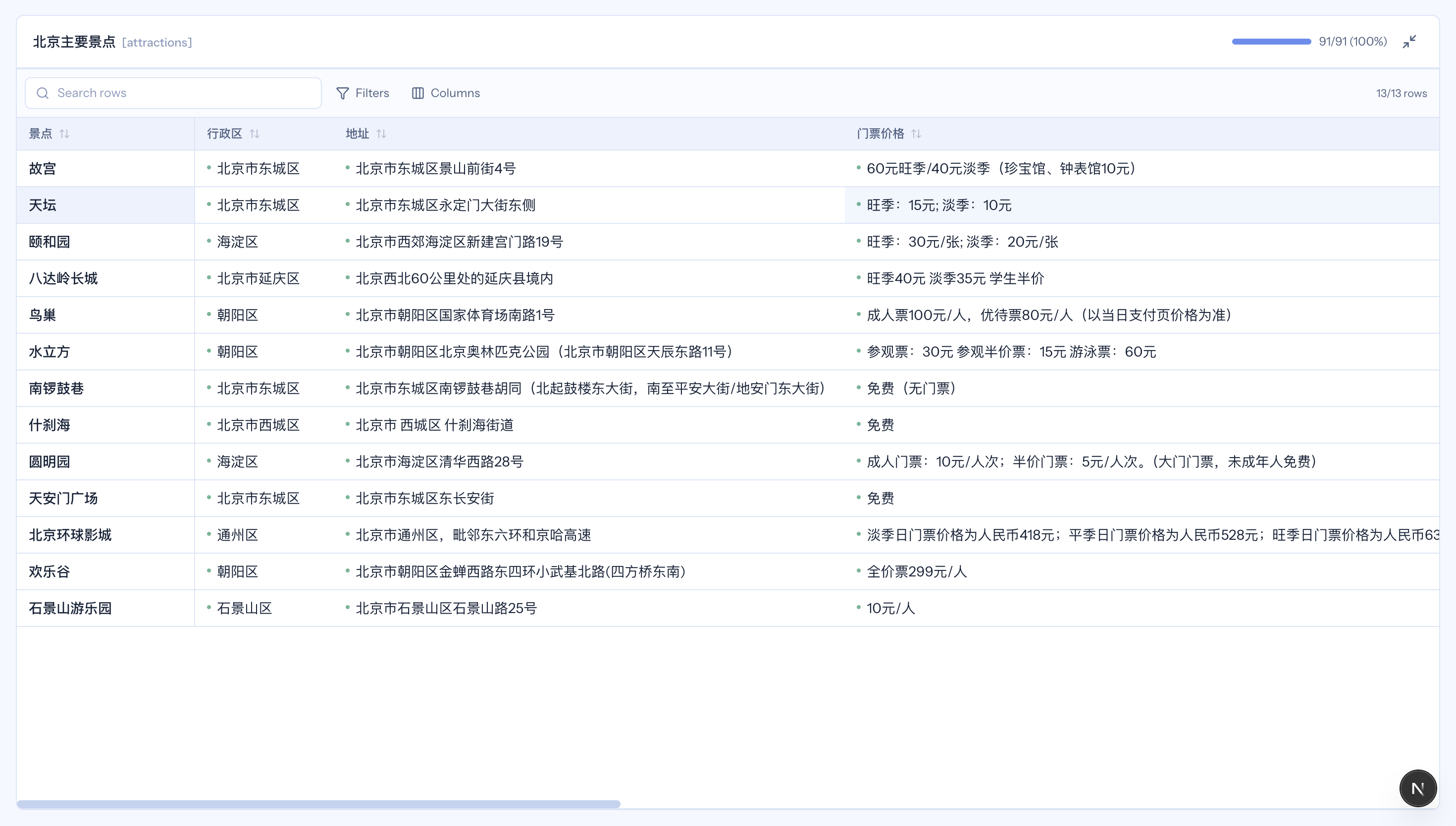}{Case 1-7}

\vspace{0.6em}
\noindent\casepanel{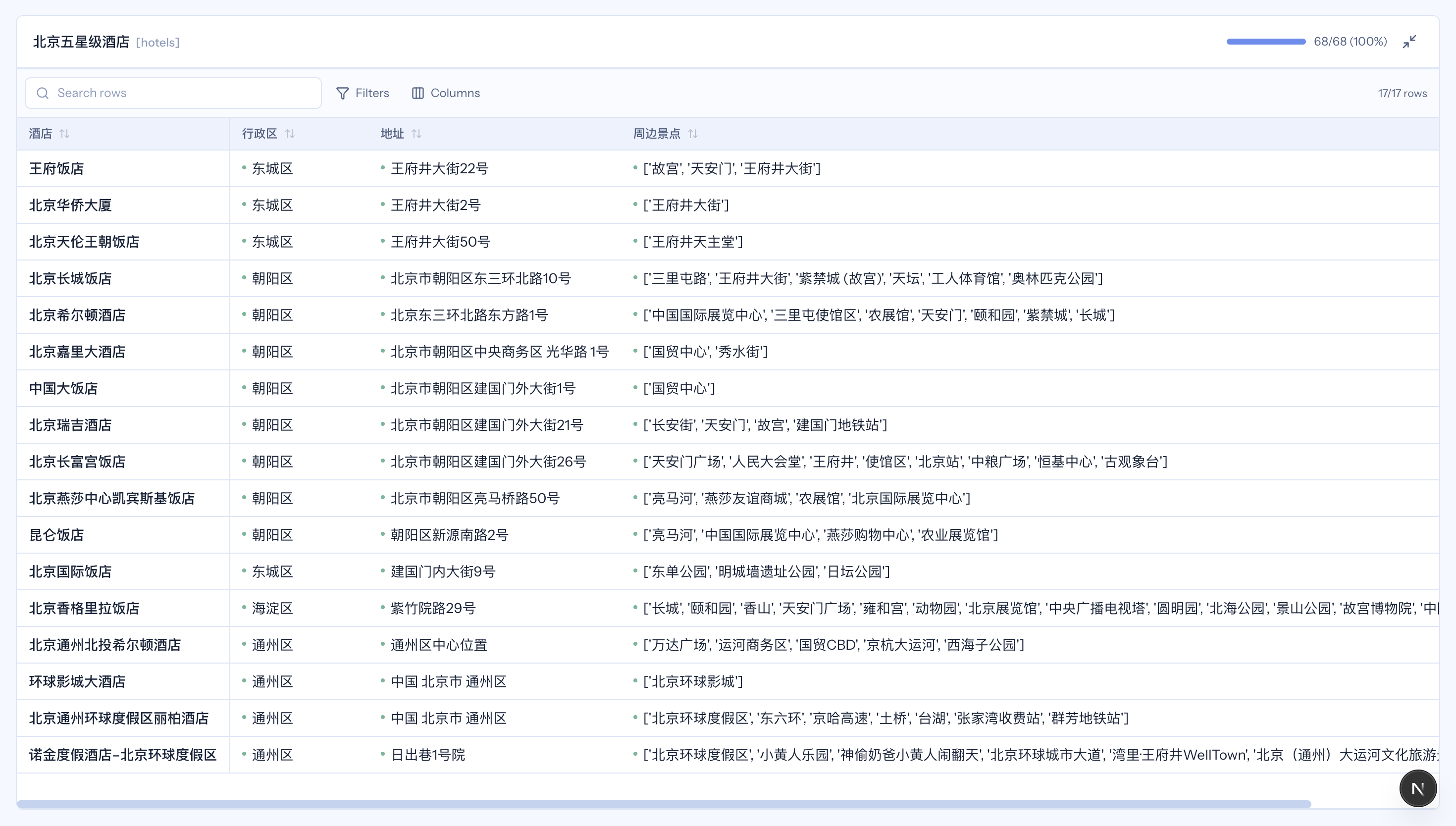}{Case 1-8}\hfill
\begin{minipage}[t]{0.485\textwidth}\mbox{}\end{minipage}

\subsection{Case 2: Agent Harness Engineering Survey}

\noindent\casepanel{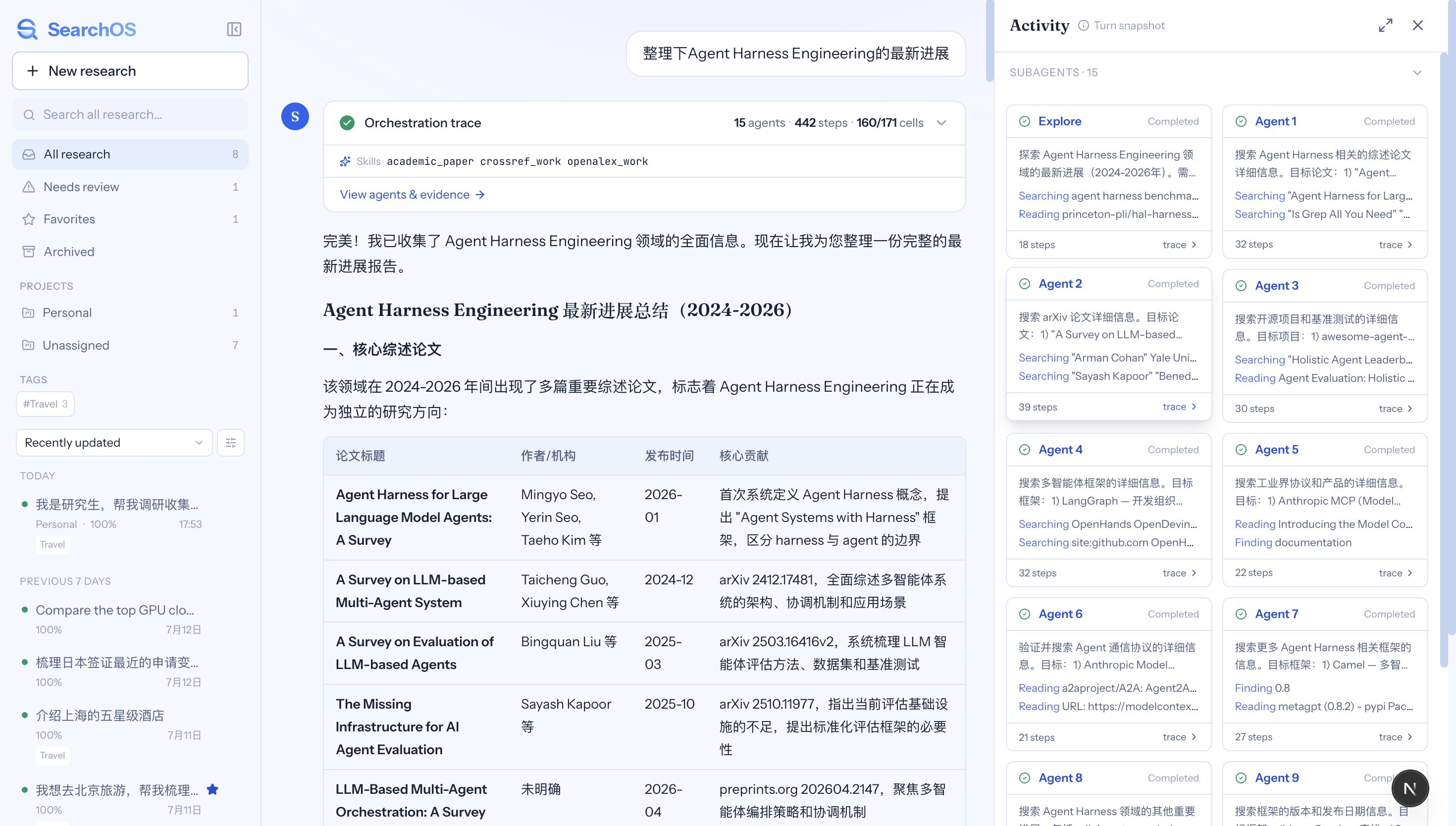}{Case 2-0}\hfill
\casepanel{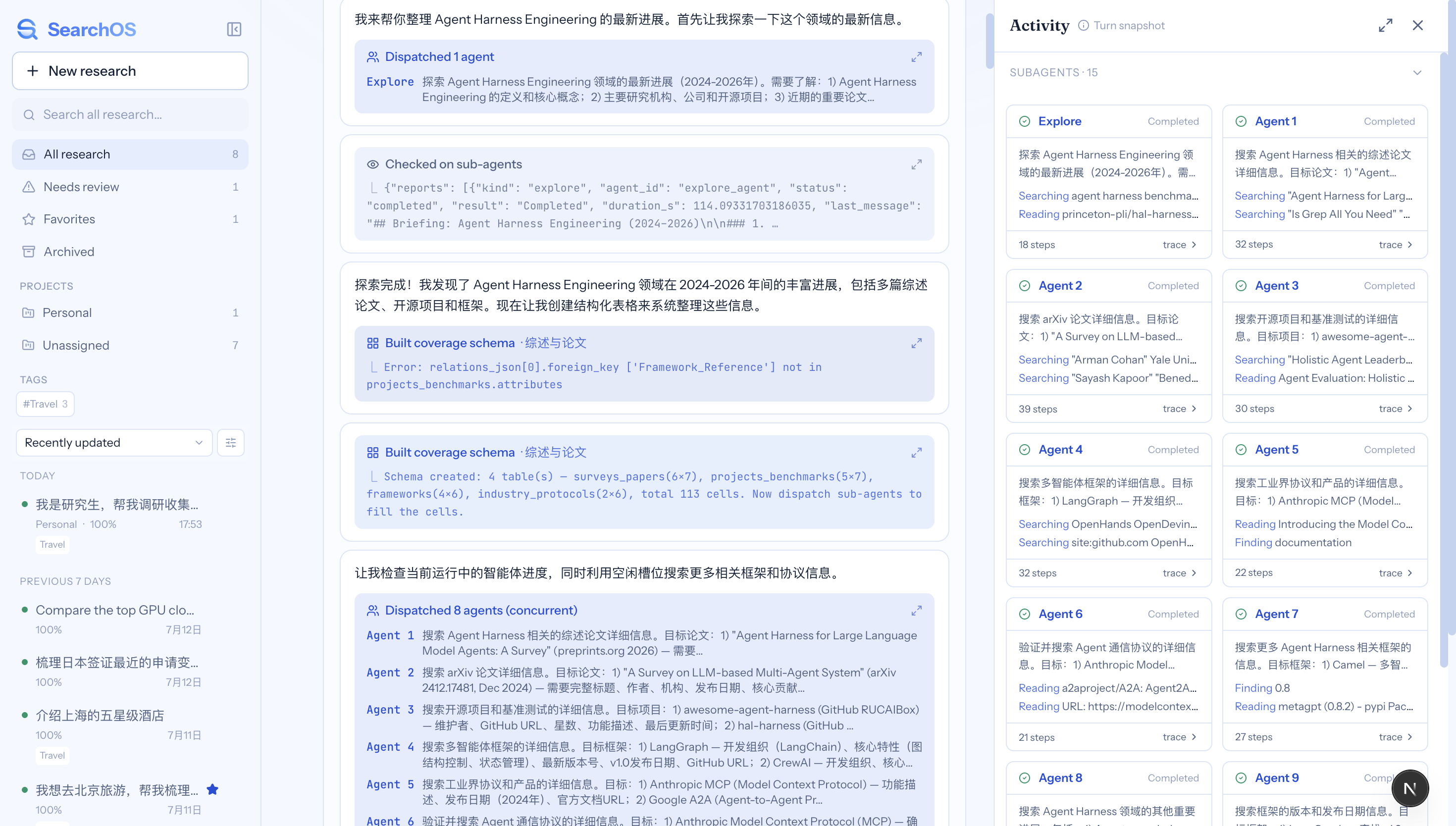}{Case 2-1}

\vspace{0.6em}
\noindent\casepanel{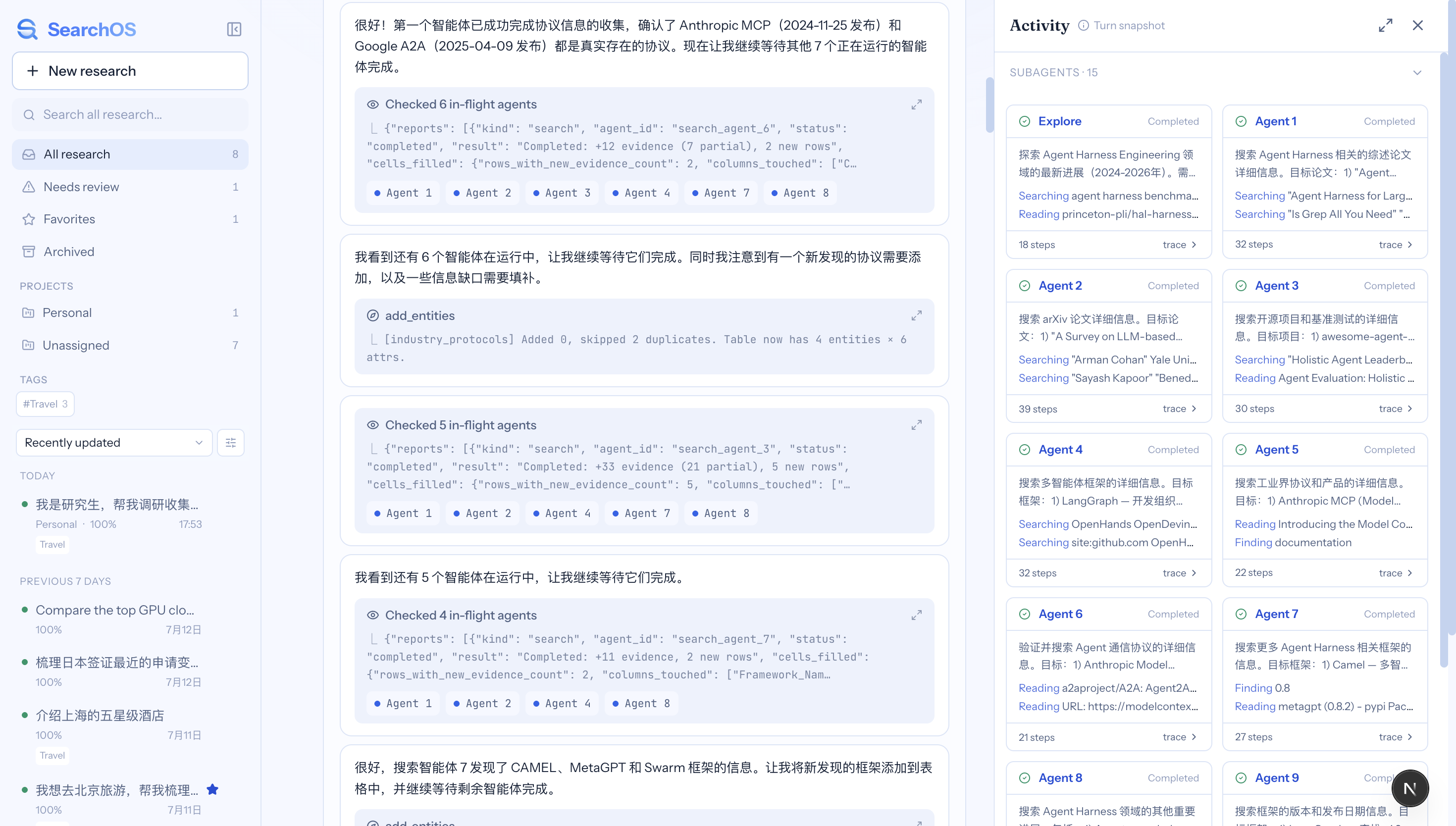}{Case 2-2}\hfill
\casepanel{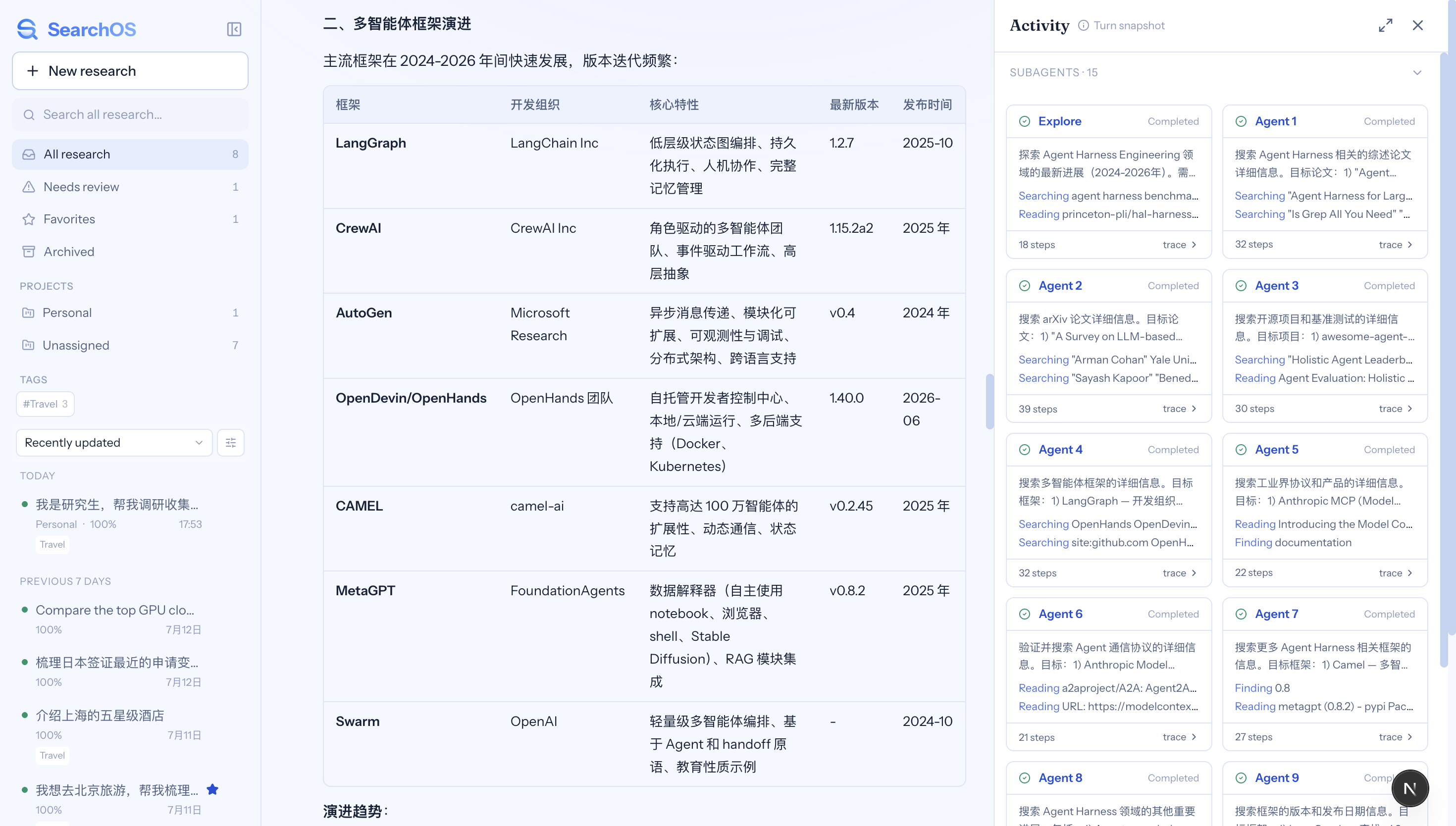}{Case 2-3}

\clearpage
\subsection{Case 3: GPU Cloud Provider Comparison}

\noindent\casepanel{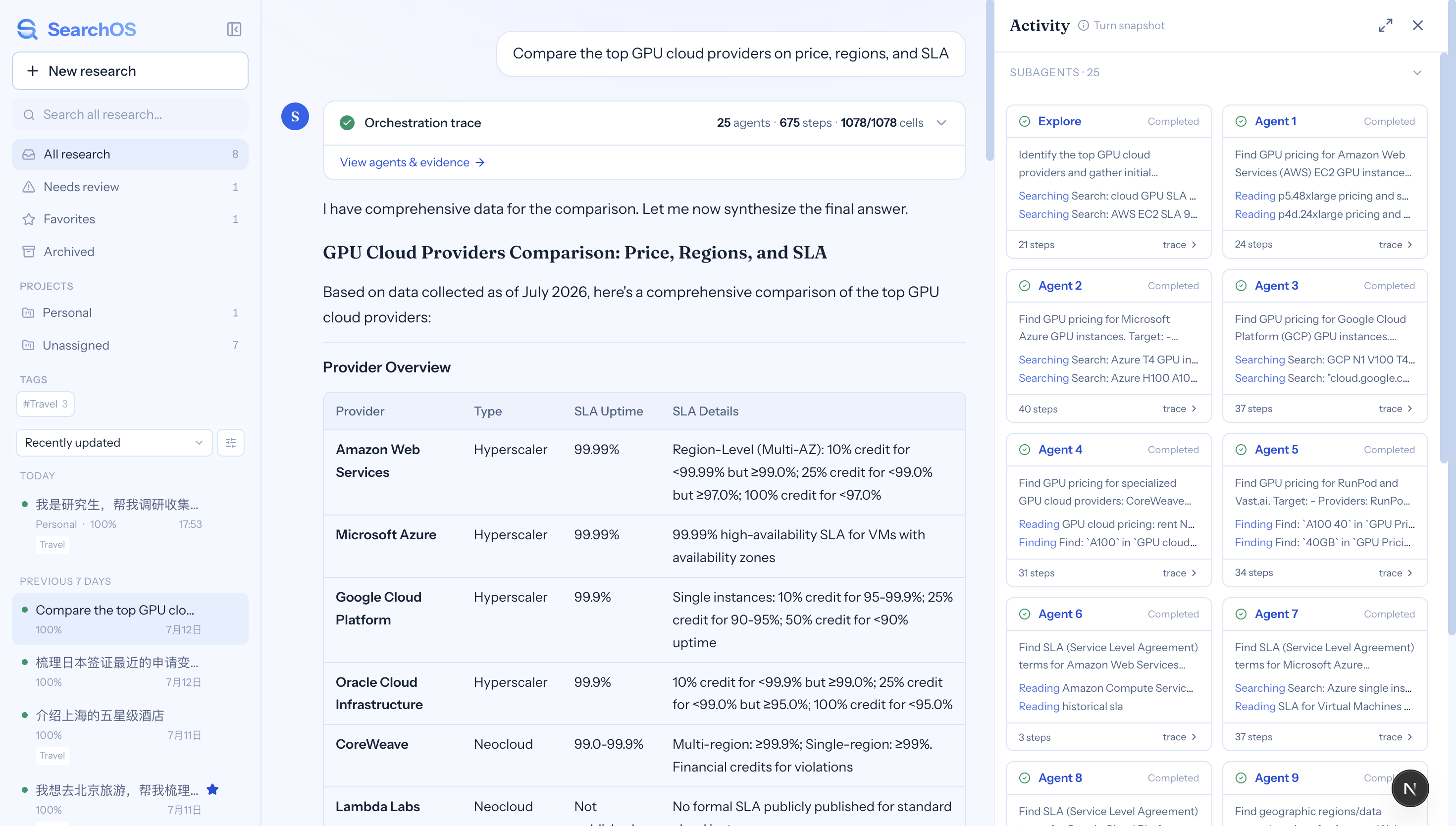}{Case 3-0}\hfill
\casepanel{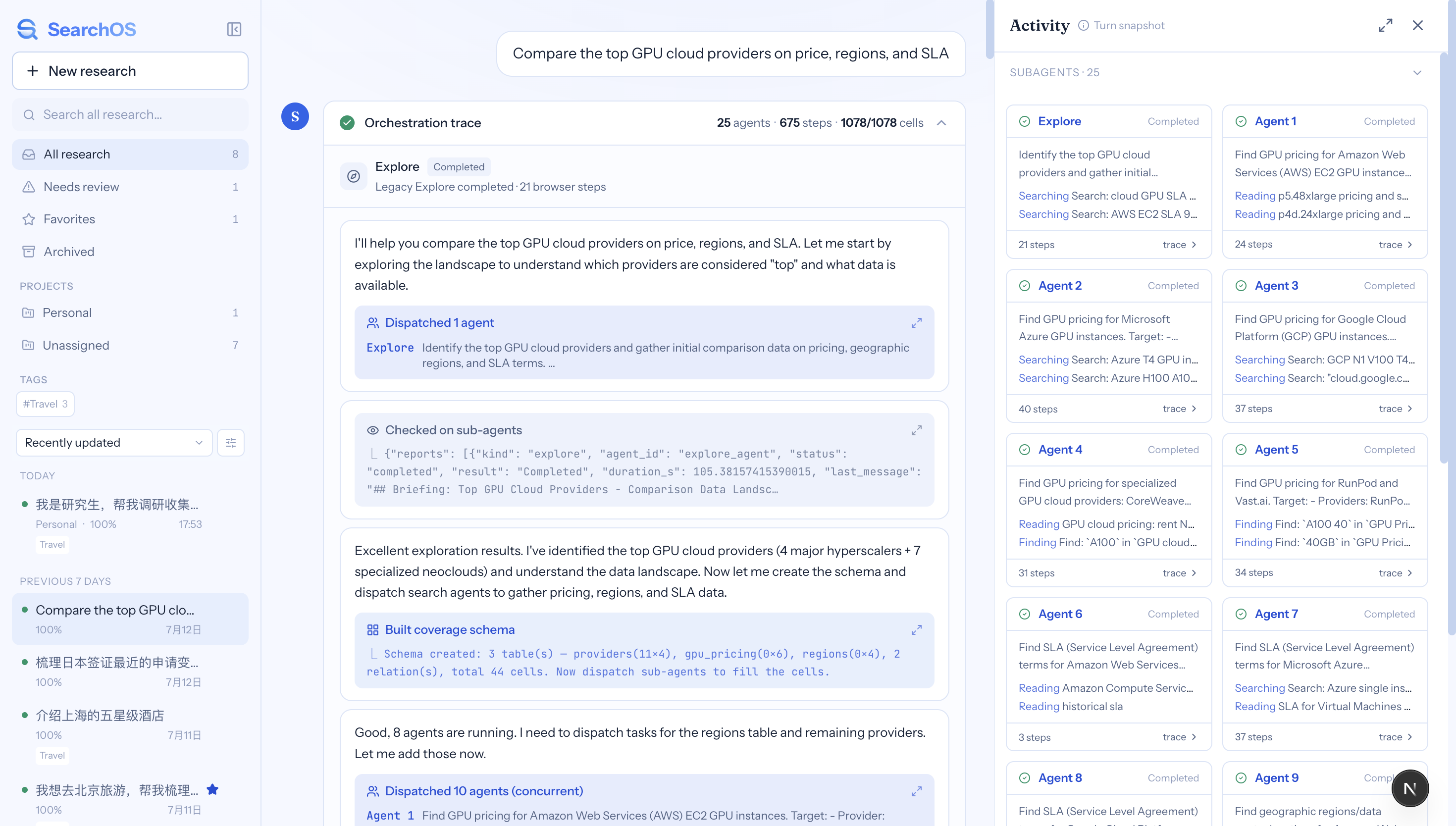}{Case 3-1}

\vspace{0.6em}
\noindent\casepanel{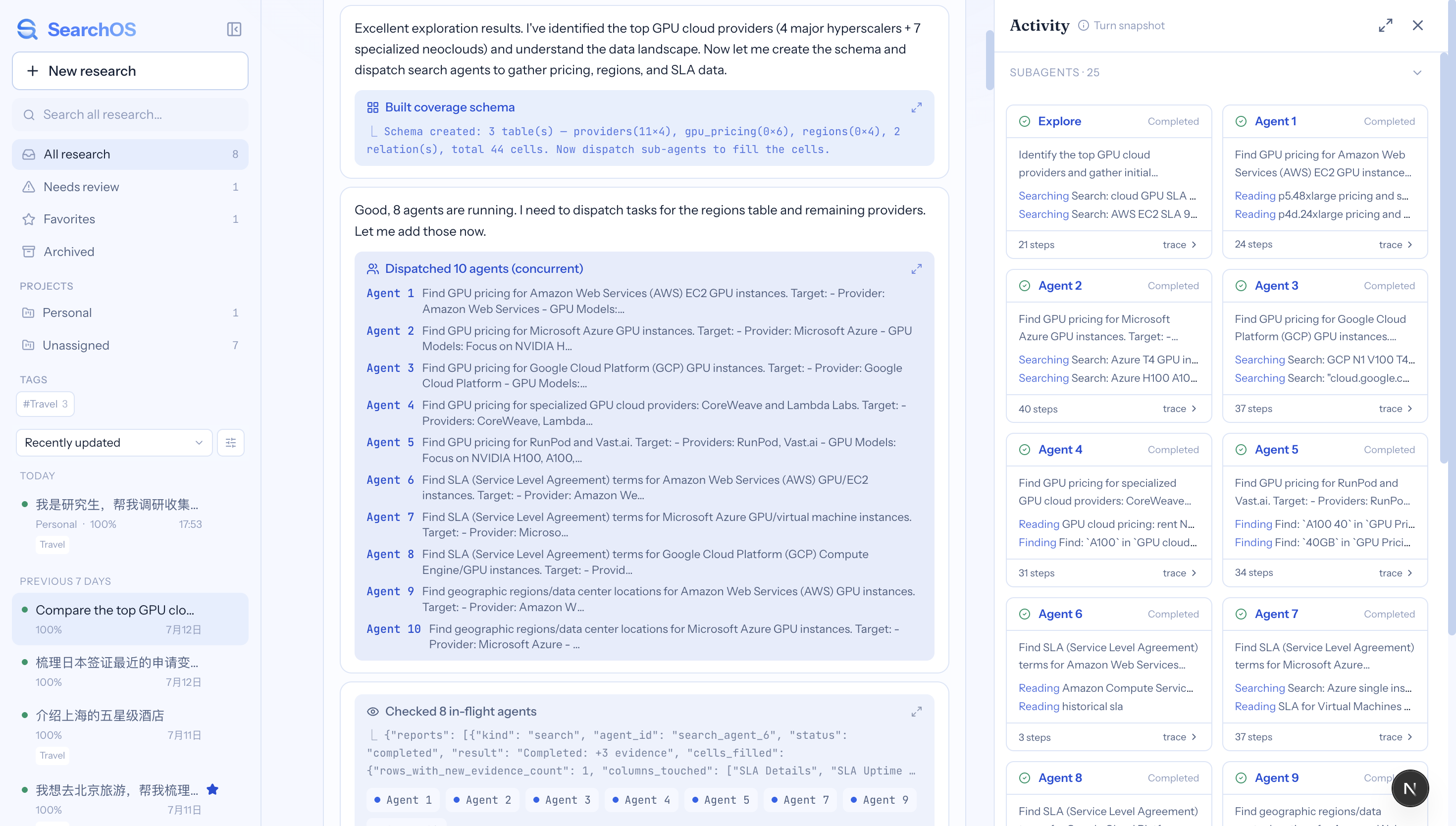}{Case 3-2}\hfill
\casepanel{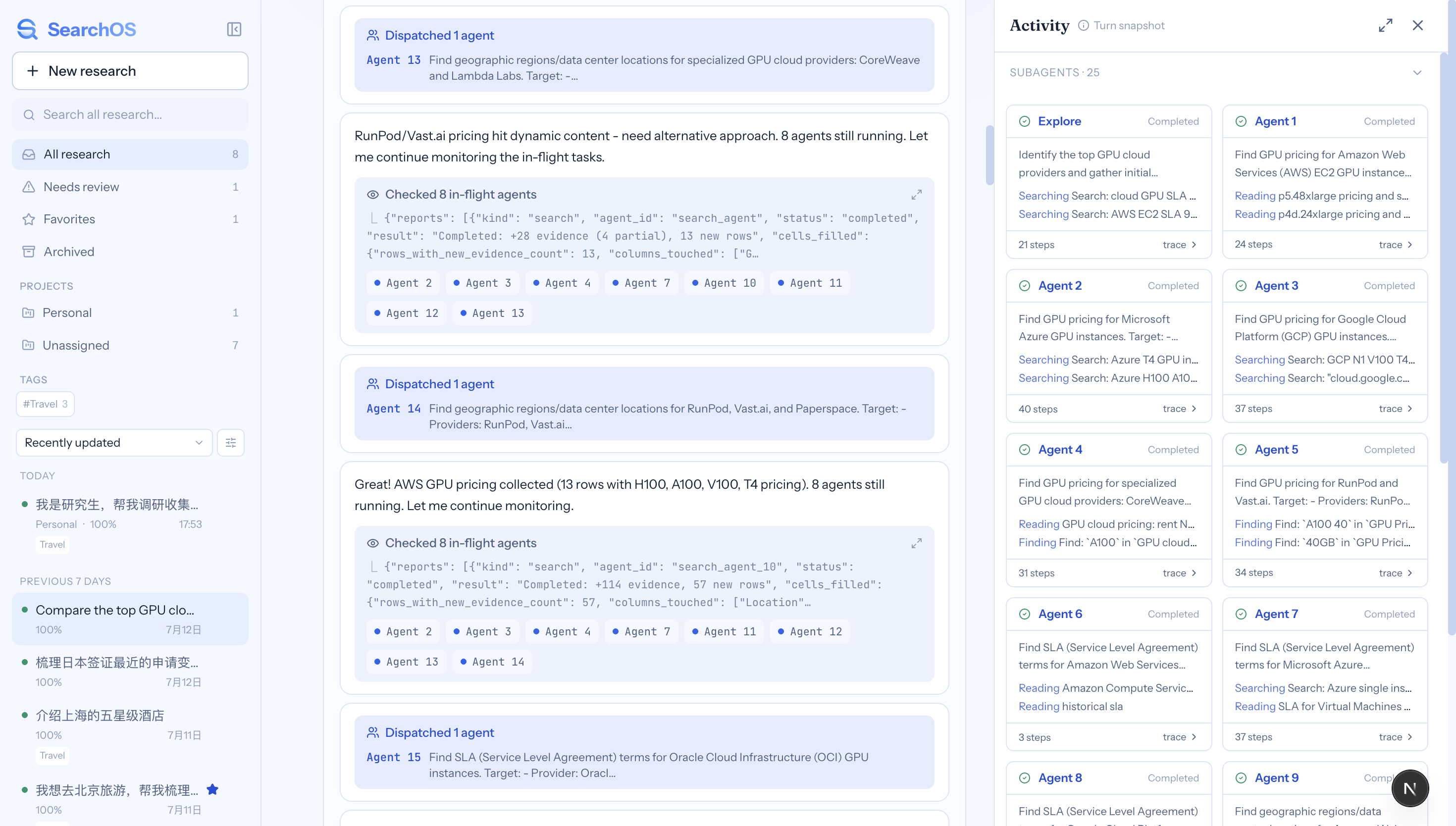}{Case 3-3}

\vspace{0.6em}
\noindent\casepanel{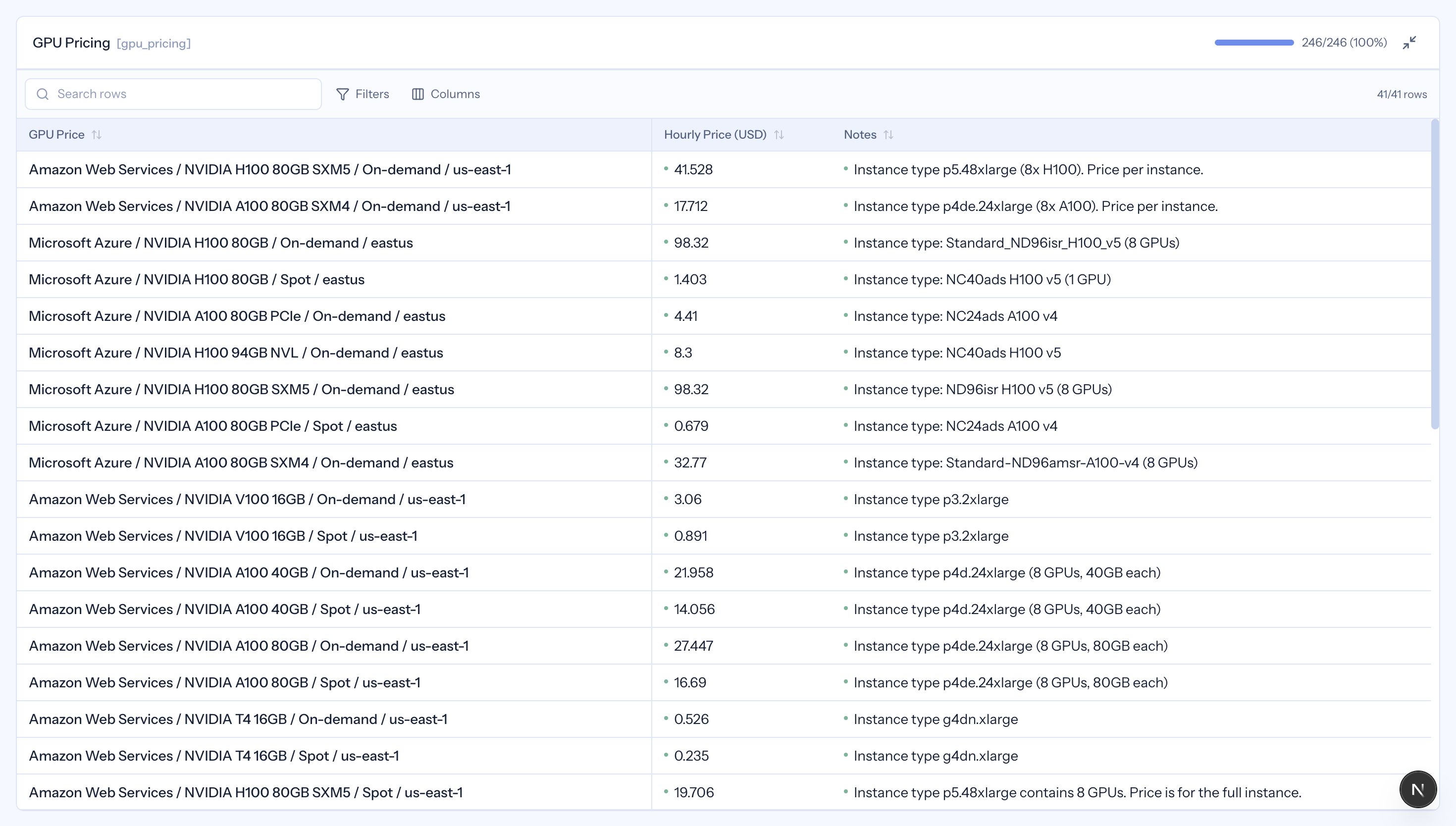}{Case 3-4}\hfill
\casepanel{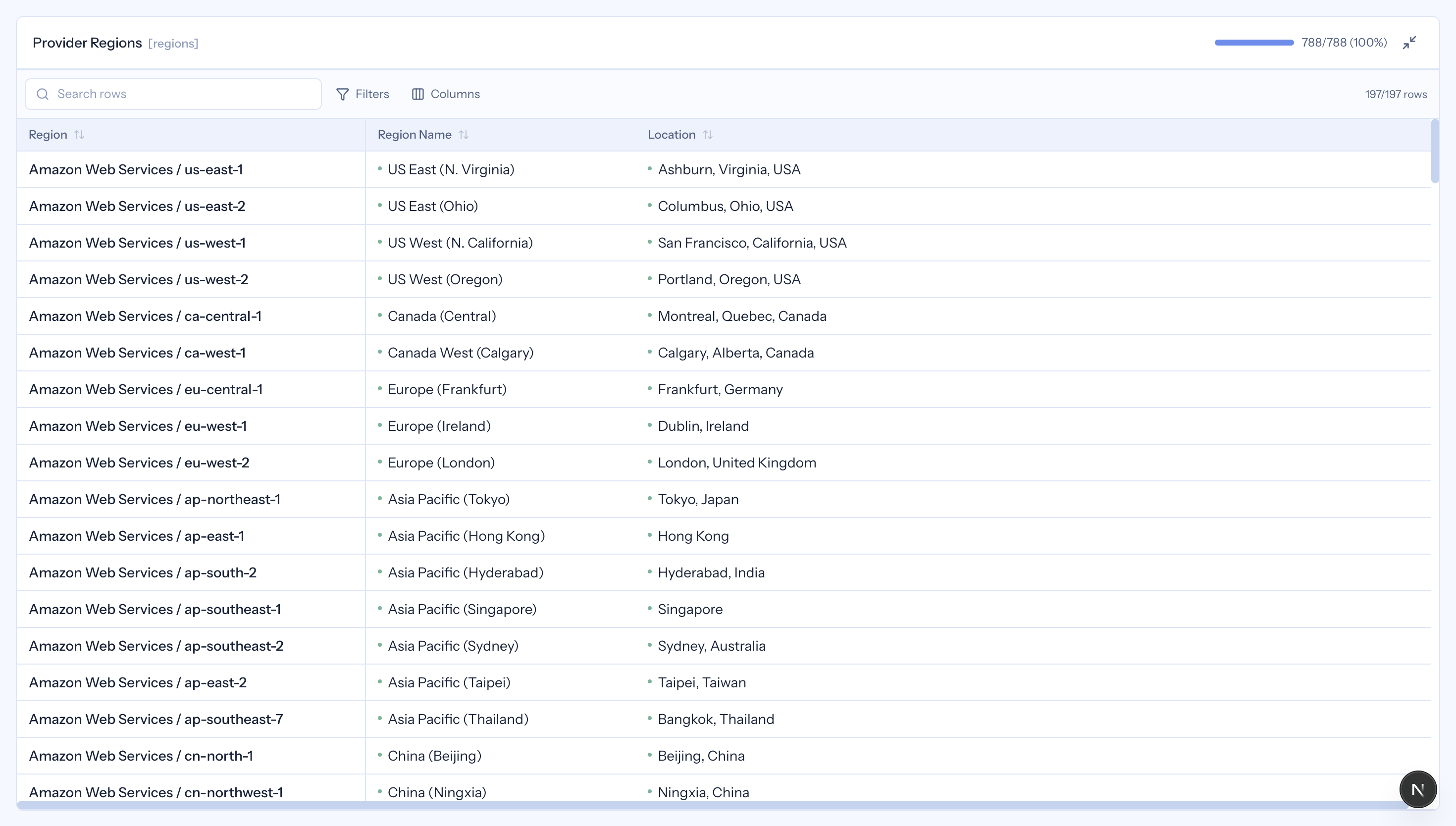}{Case 3-5}

\end{appendix}

\end{document}